\documentclass[10pt,journal,compsoc]{IEEEtran}
\ifCLASSOPTIONcompsoc
  \usepackage[nocompress]{cite}
\else
  \usepackage{cite}
\fi

\ifCLASSINFOpdf
\else
\fi
\usepackage[inkscapearea=page]{svg}
\usepackage{adjustbox}
\usepackage{graphicx,lipsum}
\usepackage{adjustbox}
\usepackage{sidecap}
\usepackage{xcolor}

\usepackage{CJKutf8}

\usepackage{amsmath}
\usepackage{amssymb}

\usepackage{hyperref}
\hypersetup{
    colorlinks=true,
    linkcolor=black,
    anchorcolor=black,
    citecolor=black,
    filecolor=black,
    menucolor=black,
    runcolor=black,
    urlcolor=black
    }

\definecolor{navyblue}{rgb}{0.0, 0.0, 0.5}

    \usepackage{flushend}

\begin{document}


\def\arraystretch{1.25}


%
\title{HICEM: A High-Coverage Emotion Model for Artificial Emotional Intelligence}
%
%
%
%

\author{Benjamin~Wortman and
        James~Z.~Wang

\IEEEcompsocitemizethanks{\IEEEcompsocthanksitem B. Wortman and J. Z. Wang are with the College of Information Sciences and Technology, The Pennsylvania State University, University Park, PA 16802, USA.
(e-mails: \{bvw5145, jwang\}@psu.edu)
}
\thanks{Manuscript received June 1, 2022.}}

\IEEEtitleabstractindextext{%
\begin{abstract}
As social robots and other intelligent machines enter the home, artificial emotional intelligence (AEI) is taking center stage to address users' desire for deeper, more meaningful human-machine interaction. To accomplish such efficacious interaction,
the next-generation AEI need comprehensive human emotion models for training.
Unlike theory of emotion, which has been the historical focus in psychology, 
emotion models are a descriptive tools. 
In practice, 
the strongest models need robust coverage, which means defining the smallest core set of emotions from which all others can be derived.
To achieve the desired coverage, we turn to word embeddings
from natural language processing. Using unsupervised clustering techniques,
our experiments show that with as few as 15 discrete emotion categories, we can provide maximum coverage across six major languages--Arabic, Chinese, English, French, Spanish, and Russian. 
In support of our findings, we also examine annotations from two large-scale emotion recognition datasets
to assess the validity of existing emotion models compared to human perception at scale. Because robust, comprehensive emotion models are foundational for developing real-world affective computing applications, this work has broad implications in social robotics, human-machine interaction, mental healthcare, and computational psychology.  
\end{abstract}

}

\maketitle

\IEEEdisplaynontitleabstractindextext

%
\IEEEpeerreviewmaketitle

\IEEEraisesectionheading{\section{Introduction}\label{sec:introduction}}


\IEEEPARstart{A}{s} far back as Darwin, researchers have studied the subjectivity and universality of human emotions~\cite{darwin_expression_2015}. 
While such research was primarily limited to academic discussions in university psychology departments, with
the rise of in-home social robots and other intelligent machines ({\it e.g.}, Alexa, Astro), has expanded this subject matter
into the field of affective computing where developing an accurate model of human emotion is a stepping stone toward artificial emotional intelligence (AEI)~\cite{krakovsky_artificial_2018}. Here, emotion modeling is a descriptive tool used to ensure that systems being developed have sufficient coverage for a wide range of human-machine or human-robot interactions.

Ideally, a robust model with sufficient coverage means
{\it identifying the smallest core set of independent human emotions from which all other emotions can be derived}. If the emotion model used in an AEI program has too many components, the AEI may struggle to distinguish among these components. On the other hand, if the model used is too simple, the AEI may not be able to understand human emotion to a level necessary for the intended application. Some researchers used the VAD dimensional model to circumvent this problem~\cite{lu2012shape}, but such an approach is not suitable for many AEI applications. 

\begin{figure*}
     \centering
     
     \begin{minipage}[b]{0.29\textwidth}
         \centering
         \textbf{Ekman}\vspace{0.1in}
         \includegraphics[width=\textwidth,
         trim = 4cm 3.5cm 8cm 2.5cm, clip]{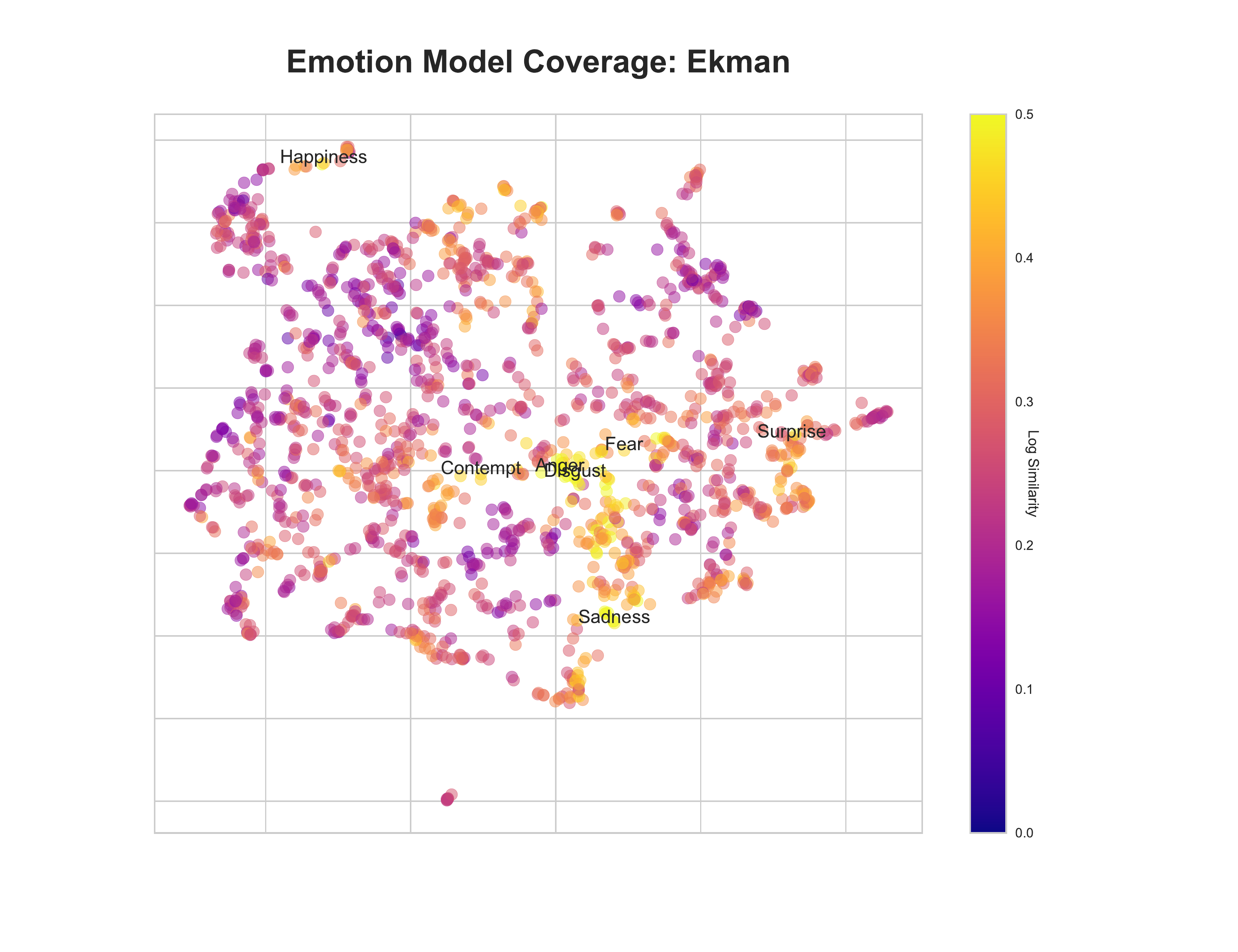}
     \end{minipage}
     \hfill
     \begin{minipage}[b]{0.29\textwidth}
         \centering
         \textbf{Plutchik}\vspace{0.1in}
         \includegraphics[width=\textwidth, 
         trim = 4cm 3.5cm 8cm 2.5cm, clip]{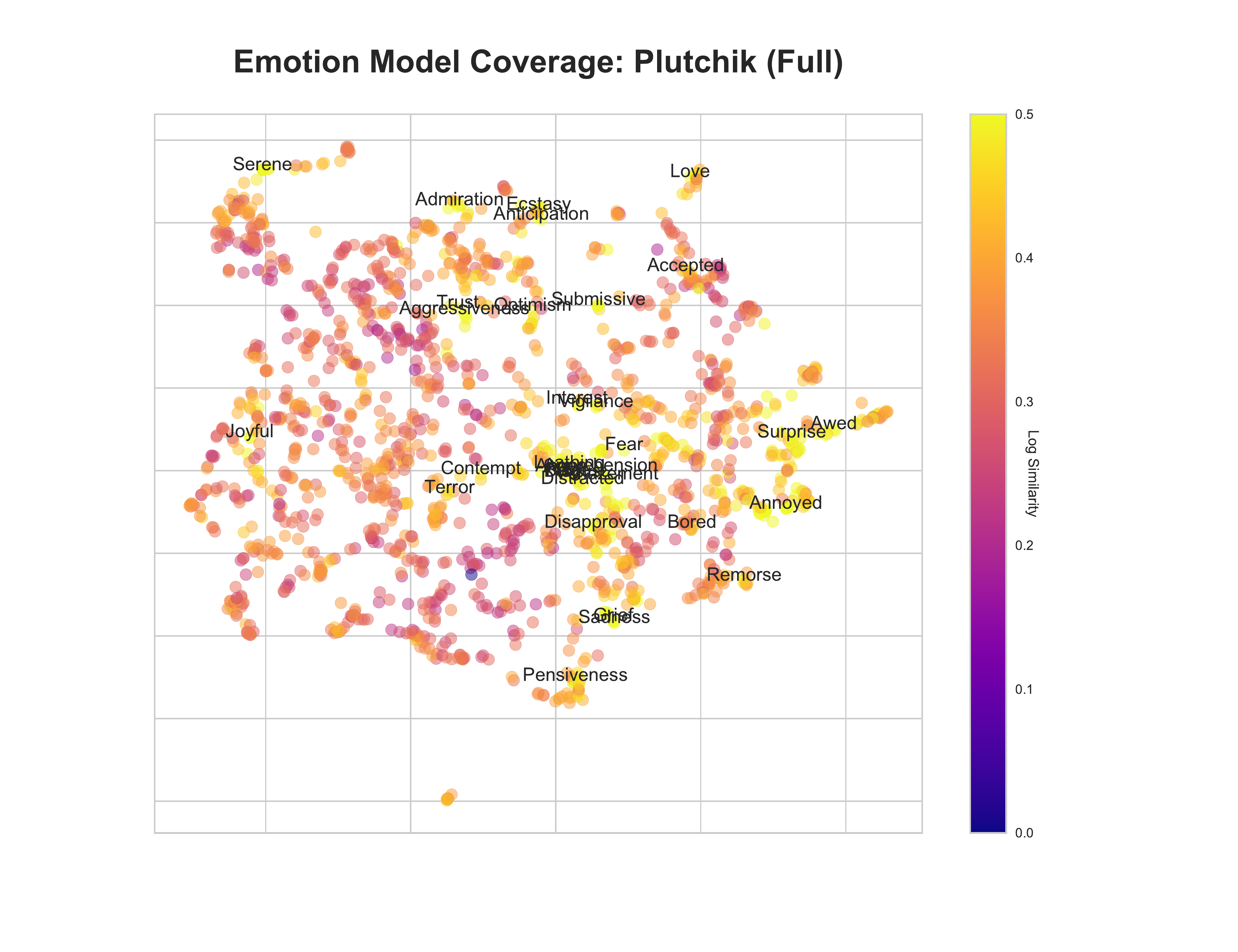}
     \end{minipage}
     \hfill
     \begin{minipage}[b]{0.29\textwidth}
         \centering
         \textbf{Cowen}\vspace{0.1in}
         \includegraphics[width=\textwidth, 
         trim = 4cm 3.5cm 8cm 2.5cm, clip]{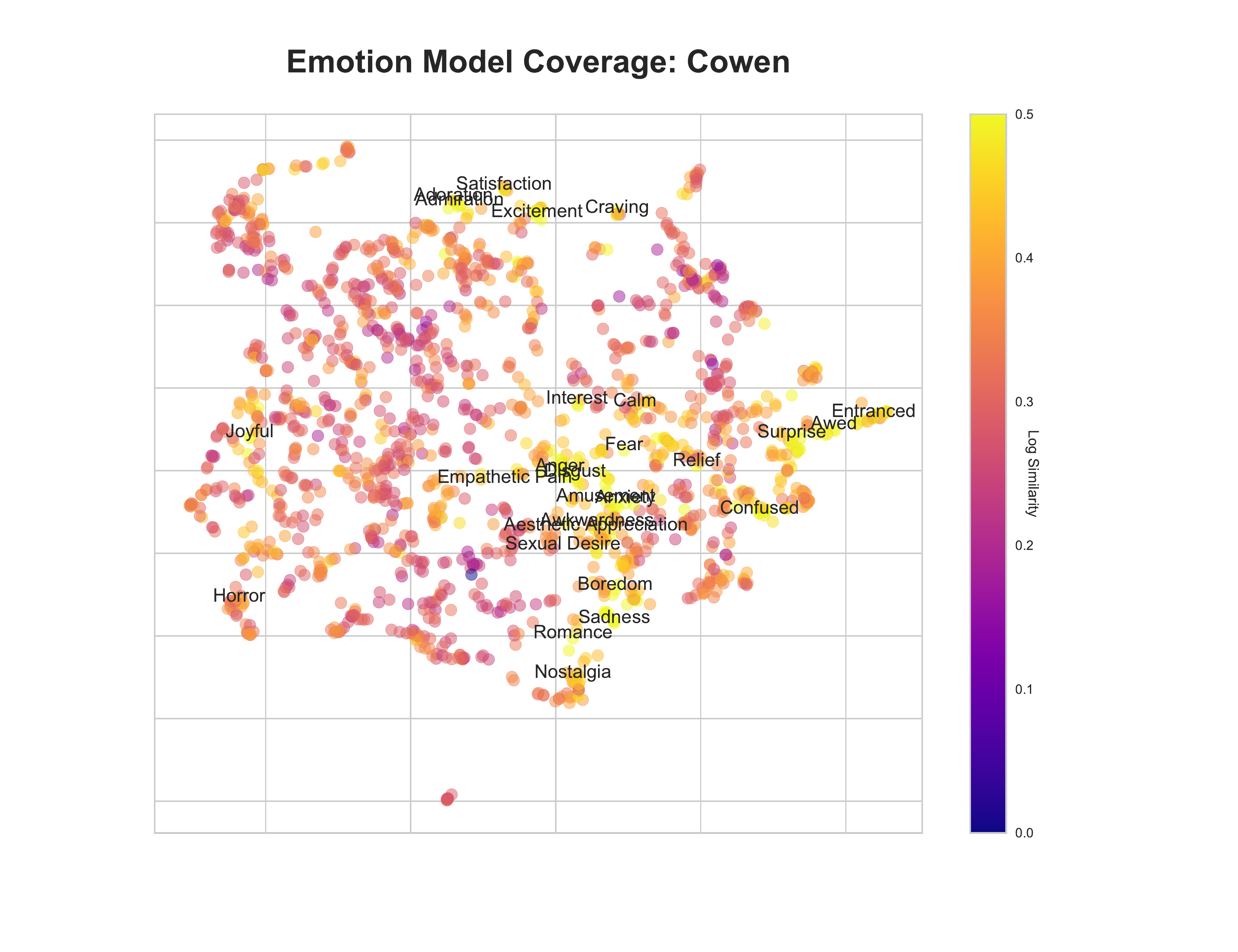}
     \end{minipage}
     \hspace{0.09\textwidth}

     \begin{minipage}[b]{0.29\textwidth}
         \centering
         \vspace{4mm}
         \textbf{GoEmotions}\vspace{0.1in}
         \includegraphics[width=\textwidth,
         trim = 4cm 3.5cm 8cm 2.5cm, clip]{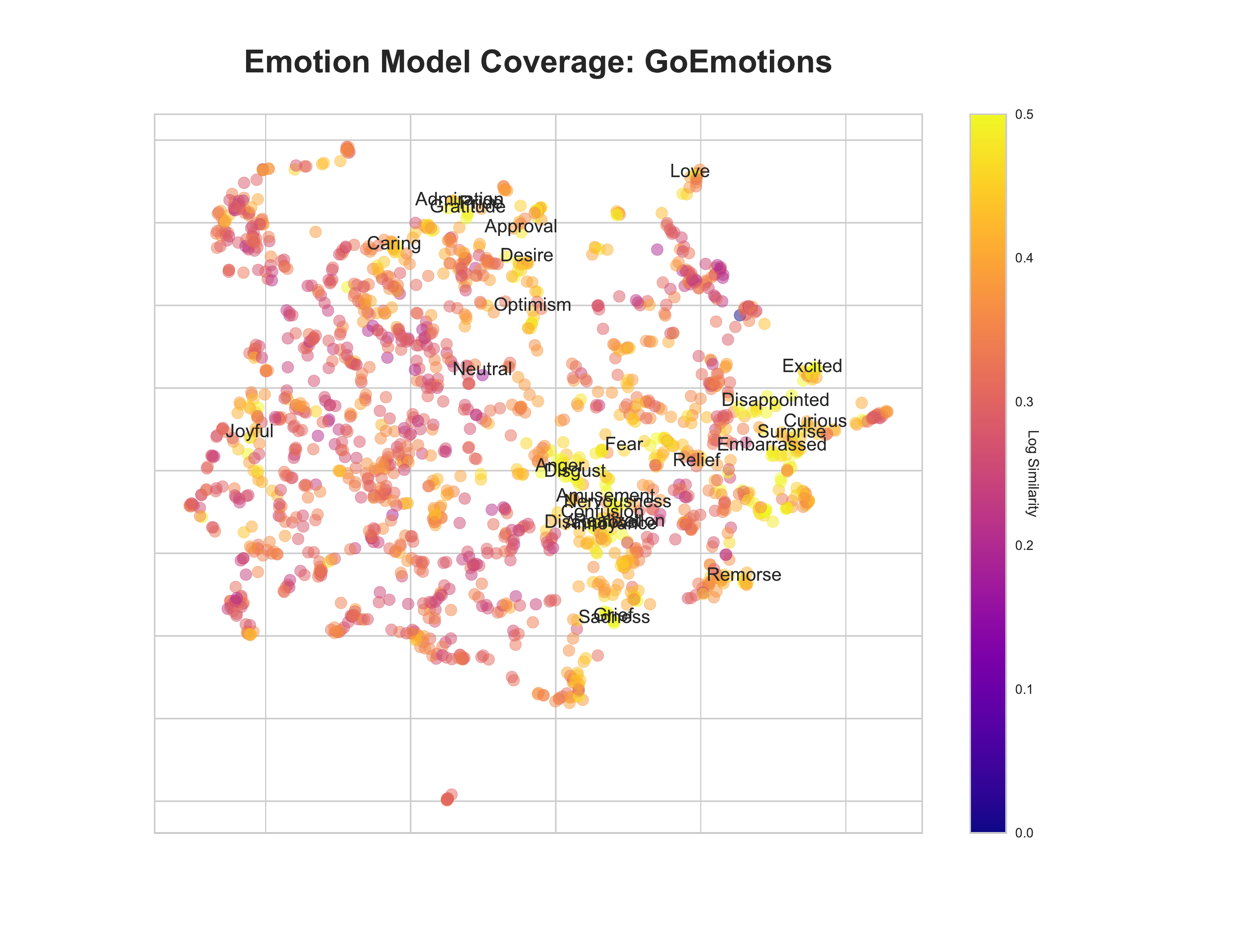}
     \end{minipage}
     \hfill
     \begin{minipage}[b]{0.29\textwidth}
         \centering
         \textbf{EMOTIC}\vspace{0.1in}
         \includegraphics[width=\textwidth, 
         trim = 4cm 3.5cm 8cm 2.5cm, clip]{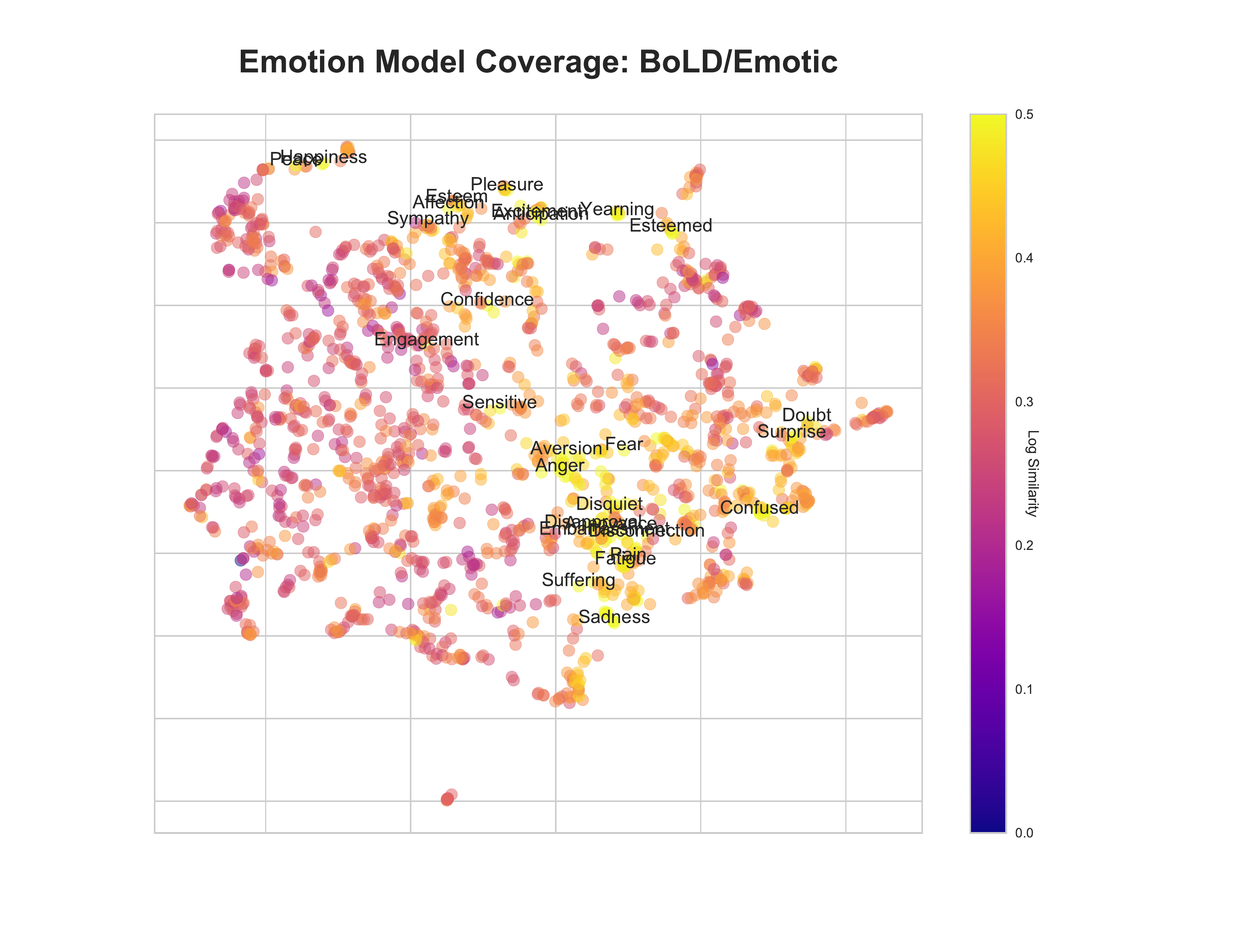}
     \end{minipage}
     \hfill
     \begin{minipage}[b]{0.29\textwidth}
         \centering
         \textbf{\textcolor{navyblue}{HICEM-15}}\vspace{0.1in}
         \includegraphics[width=\textwidth, 
         trim = 4cm 3.5cm 8cm 2.5cm, clip]{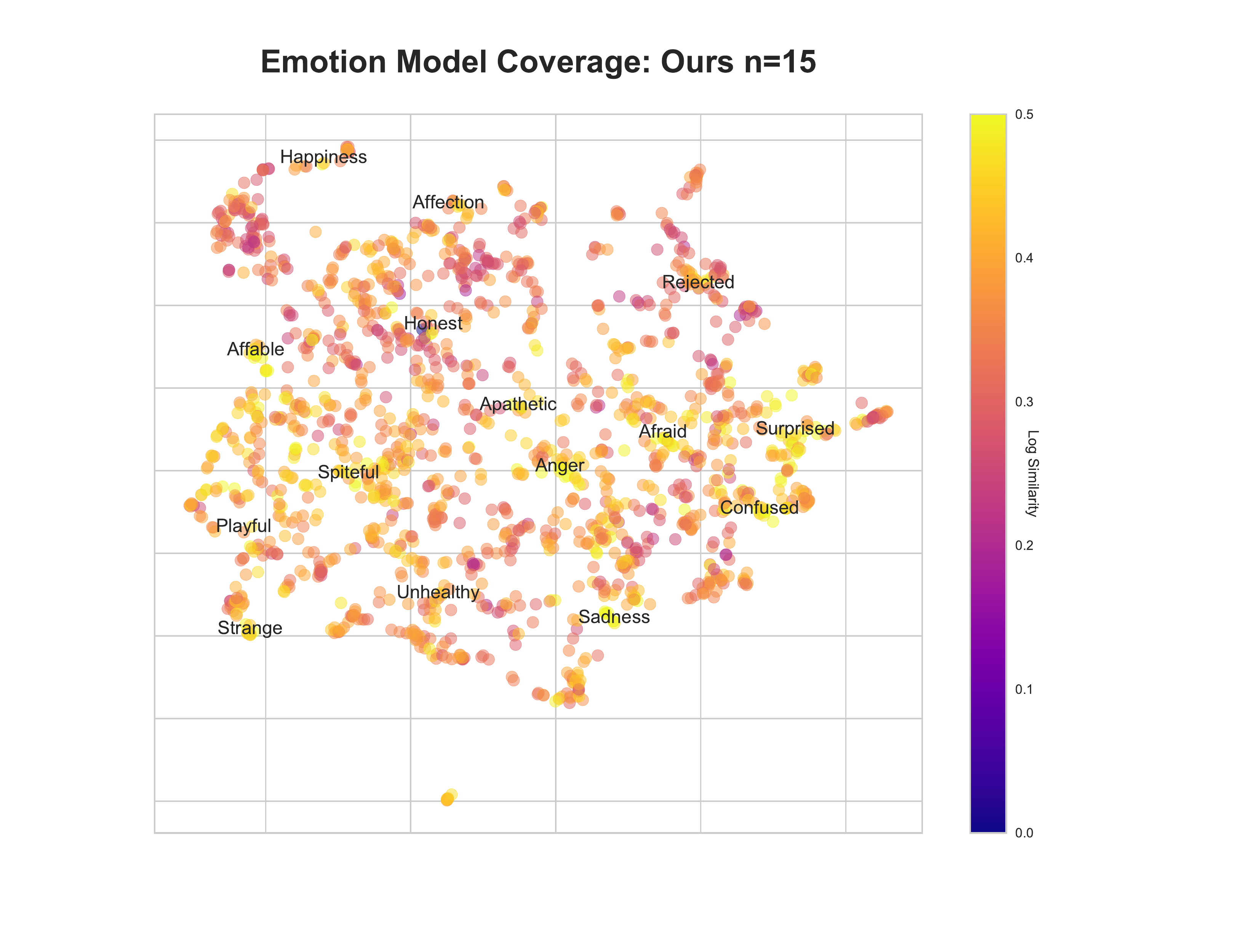}
     \end{minipage}
     \begin{minipage}[b]{0.09\textwidth}
         \centering
         \includegraphics[width=0.5\textwidth,
         trim = 0 0 0 0]{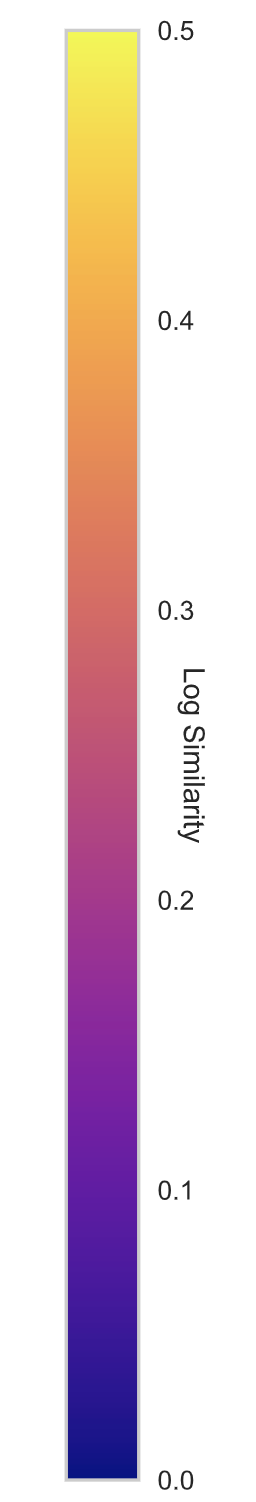}
     \end{minipage}
     

    \caption{Existing emotion models provide incomplete representations across the entire emotion space. Here we visualize the maximum log cosine similarity (with a ceiling of 0.5) between the word vectors of 1,720 emotion concepts and the contents of the model. A higher cosine similarity (yellow) means the contents of the model have a similar semantic meaning to the given concept. Let $k$ denote the number of labels or components in an emotion model. We can see Ekman's basic emotions~\cite{ekman_constants_1971} ($k$=7) are insufficient to describe the space. Plutchik's wheel of emotion~\cite{plutchik_emotions_1991} ($k$=32) is better, but its labels contain a significant amount of overlap. Using unsupervised techniques, our proposed HICEM with 15 components, named HICEM-15, minimizes the overlap between labels and provides almost the exact same coverage as Plutchik's with half as many components. For comparison we also include Cowen's emotions identified in video~\cite{cowen_self-report_2017} ($k$=27), the annotation categories for the GoEmotions Dataset~\cite{demszky_goemotions_2020} ($k$=28), and the  EMOTIC~\cite{kosti_emotic_2017} dataset annotation scheme ($k$=27).}
        \label{fig:model comparison}
\end{figure*}

Developing an ideal emotion model for AEI is a complex problem.  Existing models either do not provide enough coverage~\cite{ekman_constants_1971} or include excessive, overlapping labels to describe the space~\cite{plutchik_emotions_1991,cowen_self-report_2017,demszky_goemotions_2020}.
We provide a visual representation of an emotion model's coverage so the power of different models can be compared (Fig.~\ref{fig:model comparison}).
These are generated by taking the FastText word vectors ~\cite{grave_learning_2018} for 1,720 emotion concepts and projecting them down to two dimensions using Uniform Manifold Approximation and Projection (UMAP)~\cite{mcinnes_umap_2018}.  This dimensionality reduction technique is similar to t-SNE~\cite{hinton_stochastic_2003}, but UMAP has also been shown to help preserve global relationships. We then generate the heatmaps using the maximum log cosine similarity between the FastText word vectors for the model and our emotion-concepts list.\footnote{Since the distribution of the maximum cosine similarity across the entire list is exponential, the log of this distribution is taken to assist in visualization.} This step allows us to visualize the coverage of each emotion model in relation to our emotion concepts list. More details will be provided later.

To overcome the limitations in existing models and understand the full range of human emotion, we turn to natural language processing (NLP). Language has evolved to be the principal means of human communication and has been shown to influence our perception of the world~\cite{barrett_language_2007, lindquist_whats_2013}. By examining emotion-related words across cultures, we can identify trends and develop more robust, universal emotion models for next-generation affective computing applications. Previous work has taken similar approaches by having groups manually annotate existing emotions across continuous affective dimensions such as the valence, arousal, and dominance (VAD) space~\cite{mehrabian_pleasure-arousal-dominance_1996}. However, word embeddings popular in NLP are now able to encode this information automatically. Understanding this advancement, we leverage statistical techniques to identify the minimum number of components with maximum coverage across multiple cultures. We propose a new emotion model, named the HIgh-Coverage Emotion Model (HICEM), to provide higher coverage with fewer components compared with existing emotion models popular in psychology used for affective computing. Using two separate evaluation metrics, we show HICEM is able to achieve this goal. In support of this assertion, we also analyze the results from recent large-scale emotion recognition datasets to assess the validity and coverage of existing discrete and continuous emotion models.

The {\bf main contributions} of our work include:
\begin{itemize}
    \item We provide to the affective computing and AEI communities a new high-coverage emotion model, named HICEM, which contains more information with fewer labels than existing emotion models.
    \item We developed a principled framework to quantify the quality of existing emotion models across different languages using two new metrics and a list of 1,720 emotion concepts. We evaluated existing models with six major languages recognized by the UN.
    \item We developed a new data-driven approach, leveraging the annotations of existing large-scale emotion recognition datasets, to assess the validity of existing emotion models in relation to human perception at scale. 
\end{itemize}

The rest of the paper is organized as follows. We cover related work in emotion modeling in Section~\ref{section:related}. Section~\ref{section:cross} describes our methods for generating a high-coverage cross-cultural model of emotion, the HICEM. The methodology and analysis of existing large-scale emotion recognition datasets are shown in Section~\ref{section:largeScale}. We discuss our results and identify future areas of interest in Section~\ref{section:discussion} and conclude in Section~\ref{section:conclusion}.

\section{Related Work}\label{section:related}
There are three competing schools of thought on emotion: basic emotions, continuous models, and componential models. In this section, we briefly discuss these approaches in relation to affective computing and AEI. 

\subsection{Basic Emotion Theory}
Basic emotion theory suggests that humans evolved a set of discrete, independent emotions which when triggered produce a physiological response or action tendency. From these basic emotions, all other human emotions can be derived. As shown in Table~\ref{table:datasets_table}, these basic emotions are often used as categorical labels in affective computing datasets. More specifically, Paul Ekman's research into basic universal emotions serves as the foundation for most annotation schemes currently used~\cite{luo_arbee_2020, kosti_emotic_2017, jiang_dfew_2020, bagher_zadeh_multimodal_2018, barros_omg-emotion_2018, nojavanasghari_emoreact_2016}. His original research identified six emotions universally recognizable by their facial expression~\cite{ekman_constants_1971}. They are fear, anger, joy, sadness, disgust, and surprise.
However, several studies~\cite{ekman_facial_1993, barrett_context_2011, le_mau_professional_2021} suggest facial expressions alone are insufficient to differentiate emotions. Since it has been demonstrated that body language cues are also universal across cultures~\cite{parkinson_emotions_2017}, there may exist a subset of emotions that are universal for body language while being indistinguishable in facial expressions alone~\cite{cordaro_recognition_2020}. Although not shown to be cross-cultural, analysis by Cowen et al. on perceived emotions from vocalization~\cite{cowen_mapping_2019}, facial expressions~\cite{cowen_what_2020}, and perceived emotion from video~\cite{cowen_self-report_2017} suggests not six but more than 24 emotion categories are required to adequately map the space. However, this list was limited in that the label space was predetermined by the researchers. In attempting to develop an emotion model for text classification, Demszky et al. expanded upon Cowen et al.'s work by using user-submitted labels to augment their emotion model. These labels were then pruned and refined to generate a more annotator-friendly list of 27 emotions and a neutral category~\cite{demszky_goemotions_2020}. 

\begin{table}
\centering
\caption{Recent Emotion Recognition Datasets}
\label{table:datasets_table}
\resizebox{0.48\textwidth}{!}{\begin{tabular}{lcccc}
\hline
\begin{tabular}[c]{@{}l@{}}\textbf{Dataset}\end{tabular} & \begin{tabular}[c]{@{}c@{}}\textbf{Labeled}\\\textbf{Samples}\end{tabular} & \begin{tabular}[c]{@{}c@{}}\textbf{Categorical}\\\textbf{Emotions}\end{tabular} & \begin{tabular}[c]{@{}c@{}}\textbf{Continuous}\\\textbf{Emotions}\end{tabular} & \textbf{Year}  \\ 
\hline
BoLD~\cite{luo_arbee_2020}                                                                   & 20k                                                                        & 26\textsuperscript{†}~                                                     & VAD                                                                       & 2020           \\
DFEW~\cite{jiang_dfew_2020}                                                                   & 16k                                                                        & 7\textsuperscript{‡}                                                       & -                                                                         & 2020           \\
GoEmotions~\cite{demszky_goemotions_2020}                                                                   & 58k                                                                        & 28\textsuperscript{‡}                                                       & -                                                                         & 2020           \\
MOSEI~\cite{bagher_zadeh_multimodal_2018}                                                                  & 23k                                                                        & 6\textsuperscript{†}~                                                      & Sentiment                                                                 & 2018           \\
OMG-Emotion~\cite{barros_omg-emotion_2018}                 & 0.6k                                                                       & 7\textsuperscript{‡}                                                       & VA                                                                        & 2018           \\
Aff-Wild~\cite{kollias_deep_2018}                                                               & 0.3k                                                                       & -                                                                          & VA                                                                        & 2018           \\
EMOTIC~\cite{kosti_emotic_2017}                                                                 & 34k                                                                        & 26\textsuperscript{†}~                                                     & VAD                                                                       & 2017           \\
EmoReact~\cite{nojavanasghari_emoreact_2016}                                                               & 1.1k                                                                       & 16\textsuperscript{‡}                                                      & V                                                                         & 2016           \\ 
\hline
\multicolumn{5}{l}{\begin{tabular}[c]{@{}l@{}}\textsuperscript{†} Contains a subset of Ekman's basic emotions~ ~~\textsuperscript{‡} Ekman's basic + neutral\\ Continuous Emotion Key: (V)alence, (A)rousal, (D)ominance
\end{tabular}}                                                                                     
\end{tabular}}
\end{table}

Although Ekman's basic emotions are the most commonly used in affective computing, other models do exist. In taking an evolutionary inspired approach, Plutchik proposes an alternative to Ekman's model which consists of eight primary affective states arranged to form a wheel of emotion~\cite{plutchik_emotions_1991}. Each of these affective states has varying degrees of intensity and when combined form more complex human emotions. Although a useful tool, this model is criticized as being too simplistic and hasn't been shown to have a strong empirical foundation~\cite{smith_critiquing_2009}. Compared with Plutchik's palette theory, Jaak Panksepp took a biological approach to understanding emotion. His work pioneered the field of affective neuroscience which works to map specific regions of the brain to emotional experience~\cite{ledoux_emotion_2000, phan_functional_2002, panksepp_archaeology_2012}. In his original work, he describes seven affective systems common across mammalian brains which control specific types of behaviours and generate distinct emotional states~\cite{panksepp_affective_2004}. He describes these structures as the ``core-SELF.''\footnote{These include SEEKING (expectancy), FEAR (anxiety), RAGE (anger), LUST (sexual excitement), CARE (nurturing), PANIC/GRIEF (sadness), and PLAY (social joy).} Despite its neurological underpinnings, Panksepp's model hasn't been widely used in the affective computing community. 

\subsection{Continuous Models}
Recognizing the limits of discrete labels for human emotions, some researchers have worked to define continuous dimensions to measure a person's affective state. As shown in Table~\ref{table:datasets_table}, annotations along continuous dimensions are often used together with basic emotions. In the simplest case, such annotations simply means labeling a sample based on how positive or negative it is. This dimension is usually described as the sentiment, pleasure, or valence of the sample. Expanding beyond one dimension, the Circumplex of Affect by Russel considers arousal (relaxed vs. aroused) and valence (pleasant vs. unpleasant) as the two fundamental dimensions which together provide a mapping for the discrete emotions~\cite{russell_circumplex_1980}. 
There is strong support for the two-dimensional approach of the Affective Circumplex. These two dimensions appear across a wide range of studies~\cite{abelson_multidimensional_1962, russell_multidimensional_1985, watson_two_1999}. Similar to Panksepp's mapping of discrete emotions, there has also been a considerable amount of work mapping valence and arousal to processes in the human brain~\cite{sackeim_hemispheric_1982, anderson_dissociated_2003, jones_arousal_2003, posner_circumplex_2005}. 

For three dimensions, another popular model comes from the researchers Mehrabian et al. who described the emotion space across pleasure-displeasure, arousal-nonarousal, and dominance-submissiveness (PAD\footnote{Valence and pleasure are often used interchangeably and sometimes referred to as VAD for valence, arousal, and dominance.})~\cite{mehrabian_pleasure-arousal-dominance_1996}. This mirrors earlier work by Osgood et al., who considered the closely related concept of control instead of dominance~\cite{Osgood_May_Miron_1975}. Here, control can be thought of in terms of both the feelings of power or weakness in addition to interpersonal dominance or submission. 
With regards to the PAD model, other proposed dimensions include anticipation-expectation, anxiety-confidence, boredom-fascination, frustration-euphoria, terror-enchantment,  and intensity (how far the person is from a state of pure, cool rationality)~\cite{kort_affective_2001, fontaine_world_2007, mckeown_semaine_2010}.

\subsection{Componential Models}
In contrast with discrete basic emotions and the previously described continuous models, some work has led to componential models derived from the appraisal theory of emotion~\cite{scherer_appraisal_2001, fontaine_world_2007, grandjean_conscious_2008, marinier_computational_2009}. Under this framework, emotion is a dynamic process thought to result from a person's repeated evaluation (appraisal) of their circumstances~\cite{arnold_emotion_1960, lazarus_r_s_psychological_1966}. This perspective has an advantage over both descriptive models leveraging basic emotions and continuous dimensions because it provides an explanation for why an emotion presents itself. Since these componential models rely heavily on subjective experience~\cite{sander_systems_2005, fontaine_world_2007}, outside of lab constrained experiments~\cite{mohammadi_multi-componential_2020, meuleman_induction_2021}, they have not been widely adopted for use in affective computing.


\begin{figure*}[ht!]
    \includegraphics[width=\textwidth,
        trim = 0 80 0 90, clip]{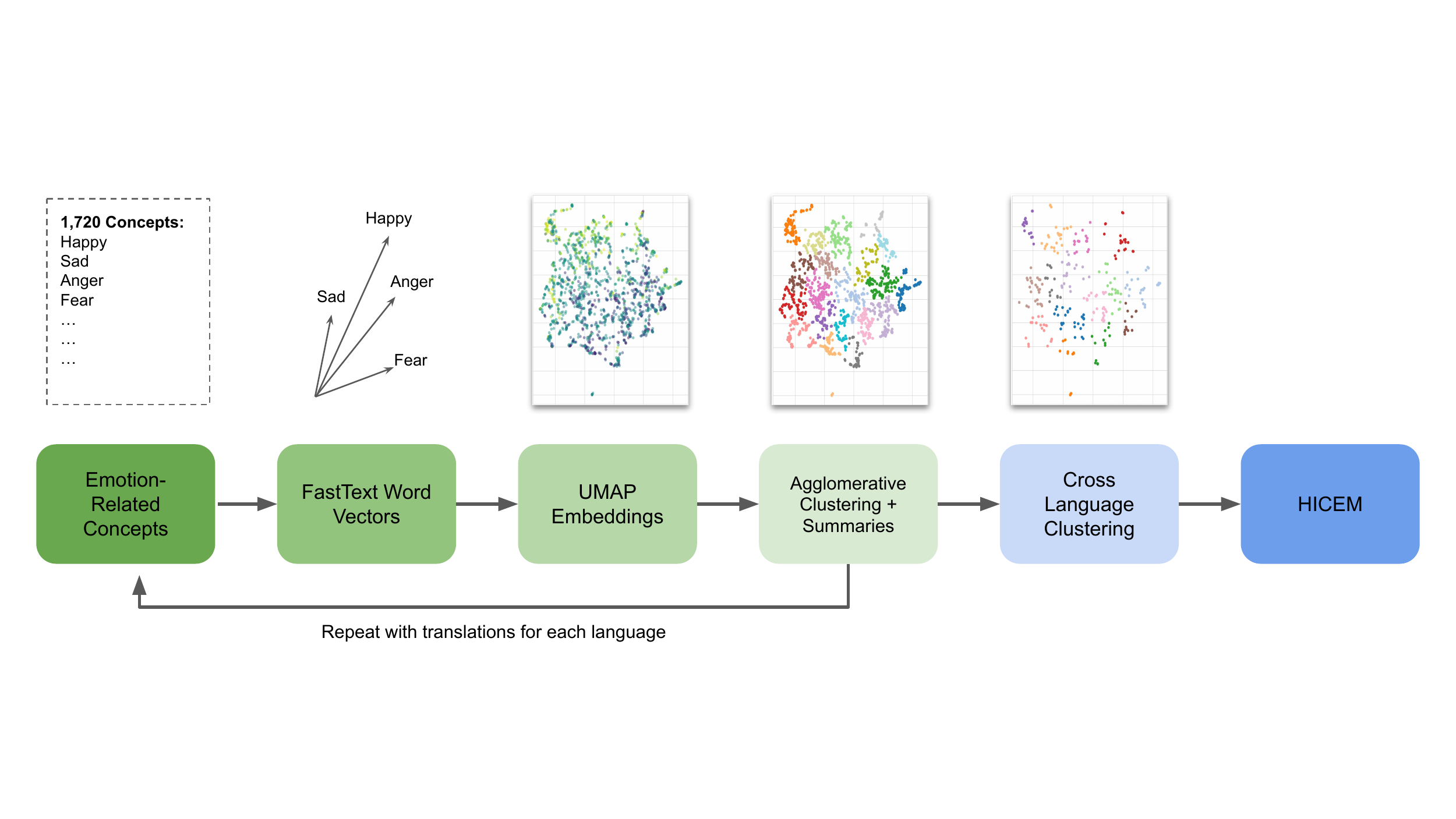}
    \caption{Our pipeline for generating a cross cultural, high-coverage model of emotion. Starting with our list of emotion-concepts, we generate their FastText word vectors. Then we perform dimensionality reduction and clustering to generate summary words. This process was repeated to create summary words for six major languages. Finally, these were once again clustered to produce our final model, the HICEM.}
    
    \label{fig:pipeline}
\end{figure*}

\section{Cross Cultural Word Embeddings}\label{section:cross}


NLP provides an interesting avenue for research into emotion modeling. By examining how current annotation schemes relate to other emotion-related words, we can take a quantitative approach toward identifying gaps in existing models. In this section, we outline our methods for generating the HIgh Coverage Emotion Model, or HICEM, from NLP word embeddings. 

\subsection{Generating a List of Emotion-Concepts}

As shown in Fig.~\ref{fig:pipeline}, we first compiled a list of emotion-related concepts from various models~\cite{ekman_constants_1971, plutchik_emotions_1991, darwin_expression_2015}, online sources~\cite{tchiki_davis_list_nodate, noauthor_380_2014,noauthor_list_2019}, and the Semantic Atlas of Emotional Concepts~\cite{averill_semantic_1975}. This list is then passed through a pre-trained Word2Vec model~\cite{mikolov_efficient_2013}, which encodes the semantic meaning of a word into a 300-dimensional vector and allows us to perform operations across these embeddings to identify relationships between words. A popular example of this process is 
$$\overrightarrow{\text{king}} - \overrightarrow{\text{man}} + \overrightarrow{\text{woman}} \approx \overrightarrow{\text{queen}}\;.$$
That is, if we take the vector for ``king,'' subtract the vector for ``man,'' and then add the vector for ``woman,'' the resulting output vector is approximately equal to ``queen.'' 
For each pairwise combination of words in this list, we append synonyms based on the cosine similarity of the average of their Word2Vec vectors. 
For example, given ``happy'' and ``sad'' we would be able to identify ``bittersweet'' because its vector is approximately the average of the two. That is,
$$\text{average}(\overrightarrow{\text{happy}},\overrightarrow{\text{sad}}) \approx \overrightarrow{\text{bittersweet}}\;.$$

This expanded set was manually pruned to remove adverbs and words unrelated to a person's emotional state. For example, words such as ``terrorist'' and ``terrorism'' are closely related to the pairwise combination ``terror'' and ``anger,'' but these have little to do with emotion so they are removed from the final expanded list. Likewise, there was a small subset of words typically used in religious contexts (`glee', `woe', `joy', `awe') which had an extremely high cosine similarity between their word vectors. This had an adverse effect on clustering so they were removed. Although, alternate forms used in different contexts remained (`gleeful', `woeful', `joyful', `awed'). Because the objective of this study was to identify gaps in existing emotion models for affective computing tasks, we took an inclusive approach and kept words that may not qualify classically as emotions but are still concepts and dispositions that influence a person's expressions~\cite{panksepp_archaeology_2012, dancy_hybrid_2021}. In total the final list contained 1,720 emotion-related keywords.\footnote{The full emotion-concept word list is available 
upon request.}

\subsection{UMAP Reduction}

Using the list of 1,720 emotion concepts, a pre-trained FastText model~\cite{joulin_bag_2016} is used to encode the semantic meaning of the word. FastText is a variant of Word2Vec that operates at the n-gram level which allows for the use of subword information to improve the quality of the embeddings~\cite{mikolov_advances_2018}. Trained on Common Crawl and Wikipedia, this model provides a 300-dimensional embedding for each word. FastText was chosen over more advanced techniques such as BERT~\cite{devlin_bert_2018} due to its location invariance and its use of the same methodology to generate vectors for multiple languages~\cite{grave_learning_2018}. BERT embeddings vary based on the location of the word within the text. In our testing, BERT produced poor results when feeding individual words to the model.

\begin{figure*}[ht!]
     \centering
     
     
     \hspace{2cm}
     \begin{minipage}[b]{0.94\textwidth}
     
         \begin{minipage}[b]{0.24\textwidth}
             \centering
             \textbf{PCA}\vspace{0.05in}
             
             \includegraphics[width=\textwidth,
             trim = 4cm 3cm 8cm 2.5cm, clip]{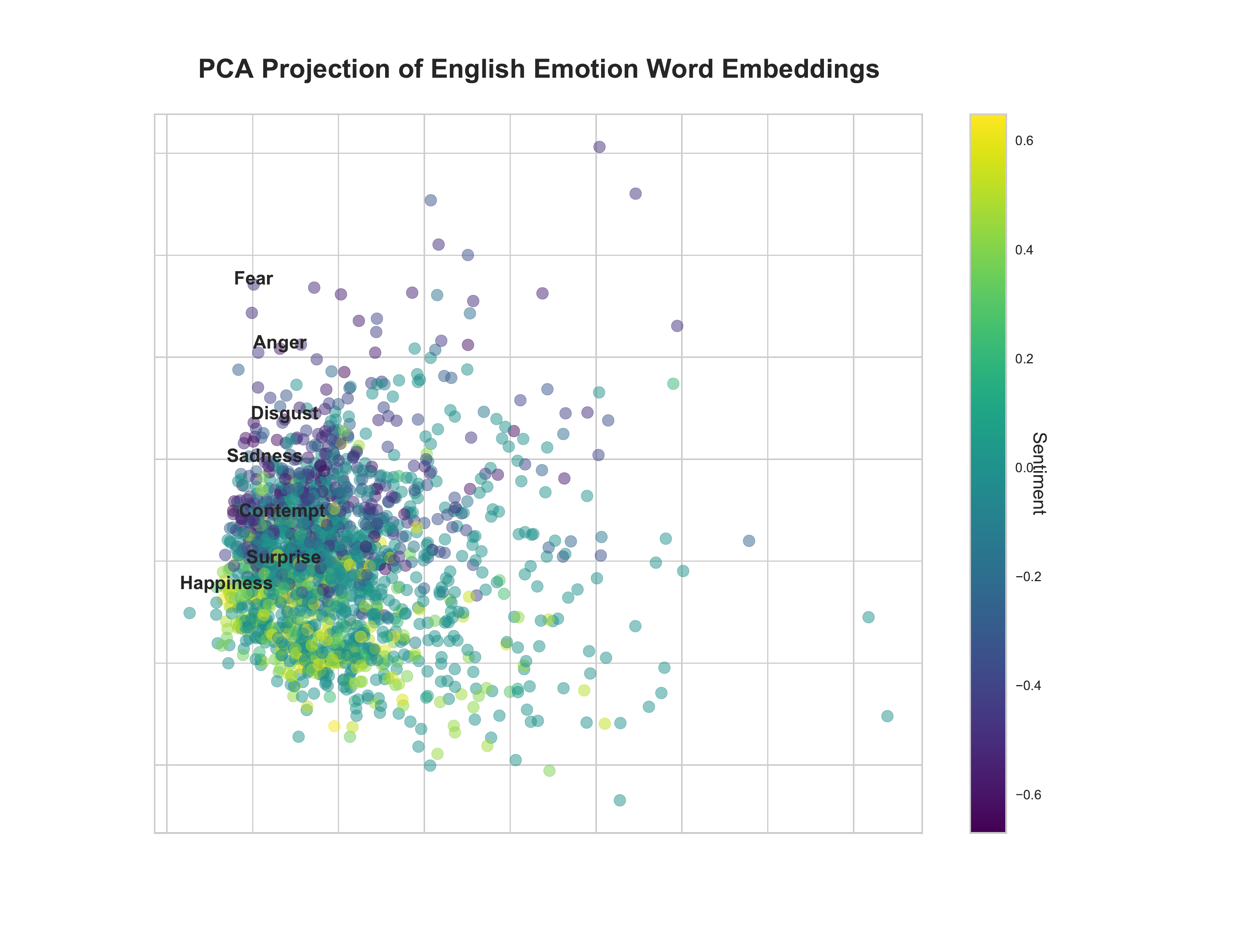}
         \end{minipage}
         \begin{minipage}[b]{0.24\textwidth}
             \centering
             \textbf{SVD}\vspace{0.05in}
             
             \includegraphics[width=\textwidth, 
             trim = 4cm 3cm 8cm 2.5cm, clip]{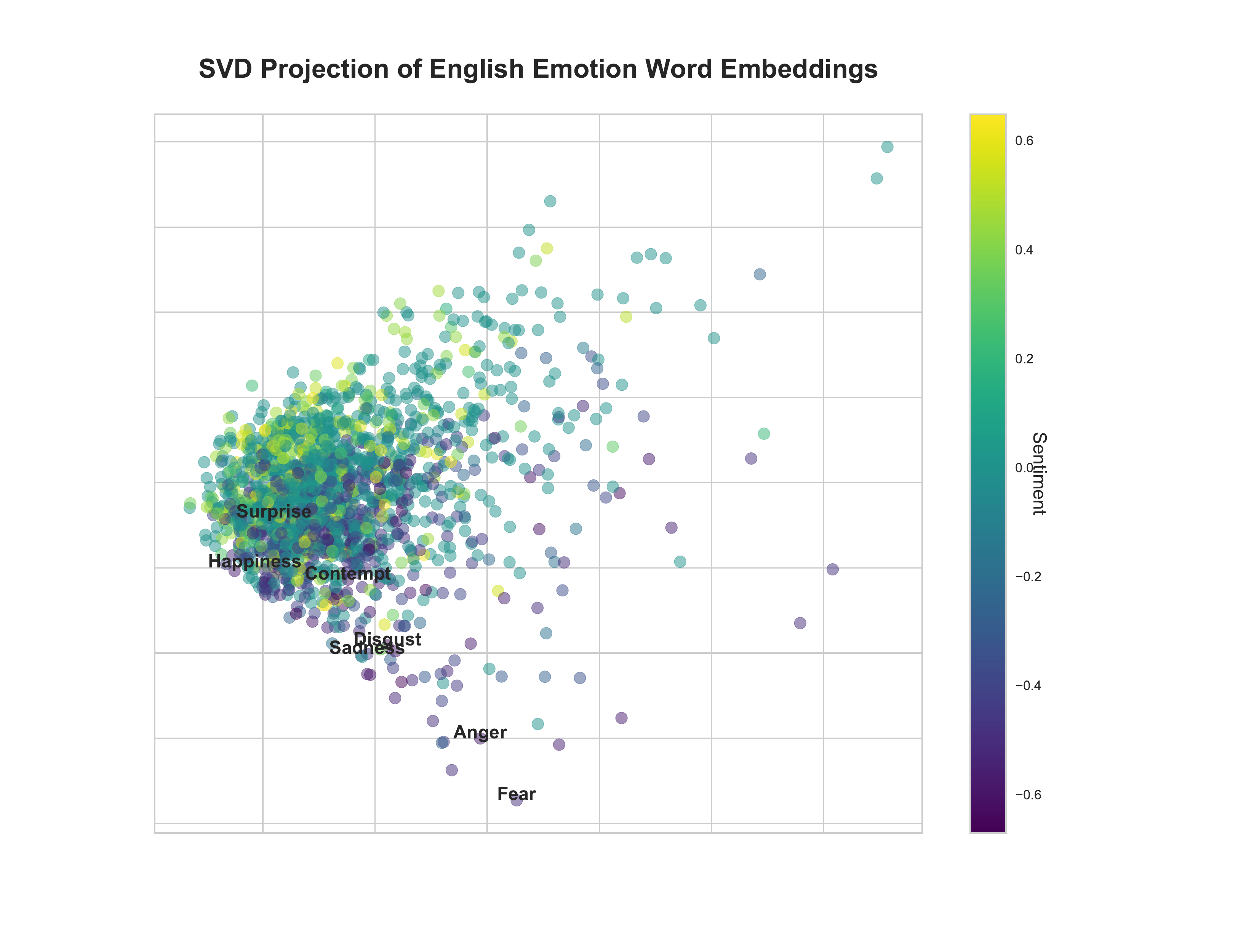}
         \end{minipage}
         \begin{minipage}[b]{0.24\textwidth}
             \centering
             \vspace{3mm}
             \textbf{t-SNE}\vspace{0.05in}
             
             \includegraphics[width=\textwidth, 
             trim = 4cm 3cm 8cm 2.5cm, clip]{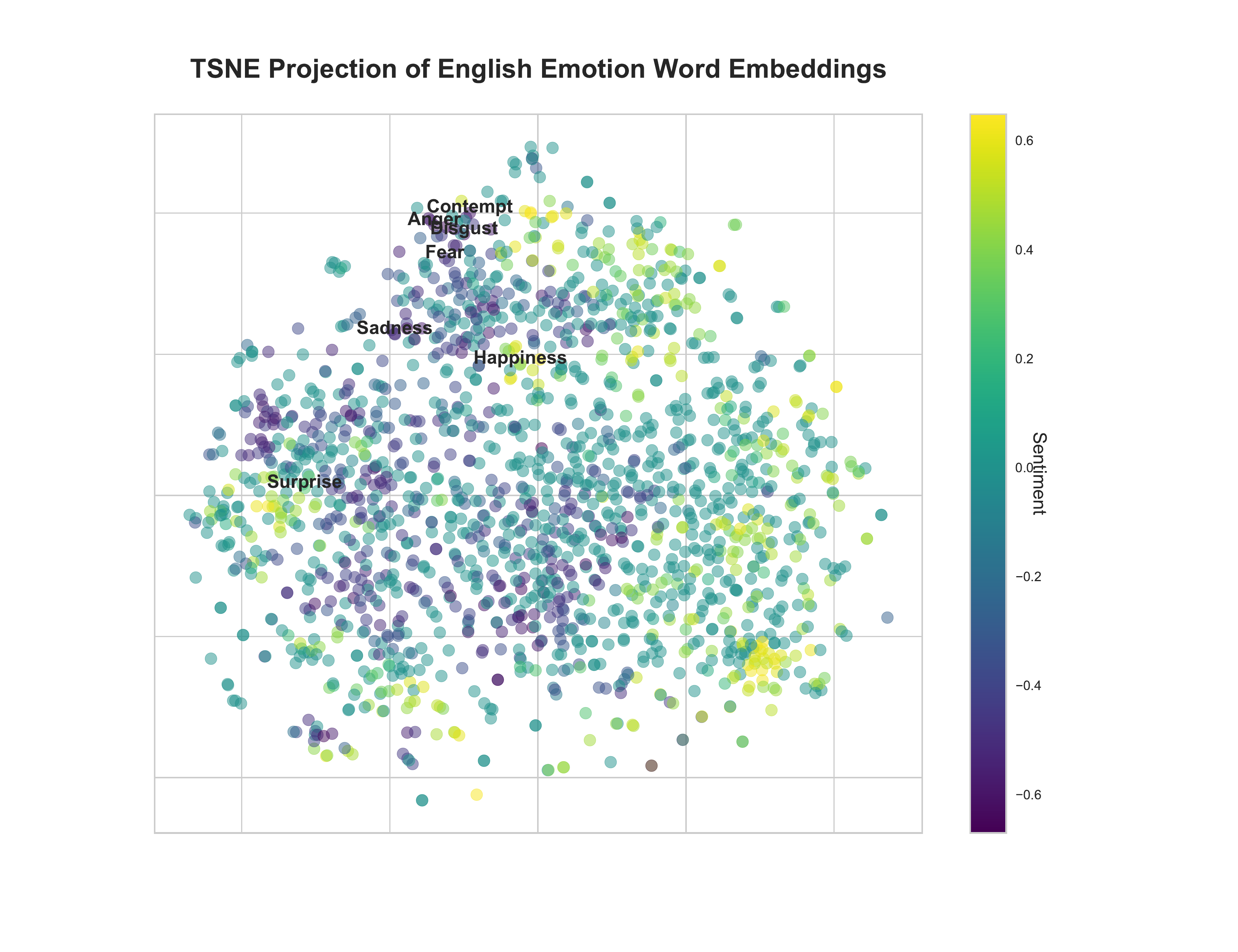}
         \end{minipage}
         \begin{minipage}[b]{0.24\textwidth}
             \centering
             \vspace{3mm}
             \textbf{UMAP}\vspace{0.05in}
             
             \includegraphics[width=\textwidth, 
             trim = 4cm 3cm 8cm 2.5cm, clip]{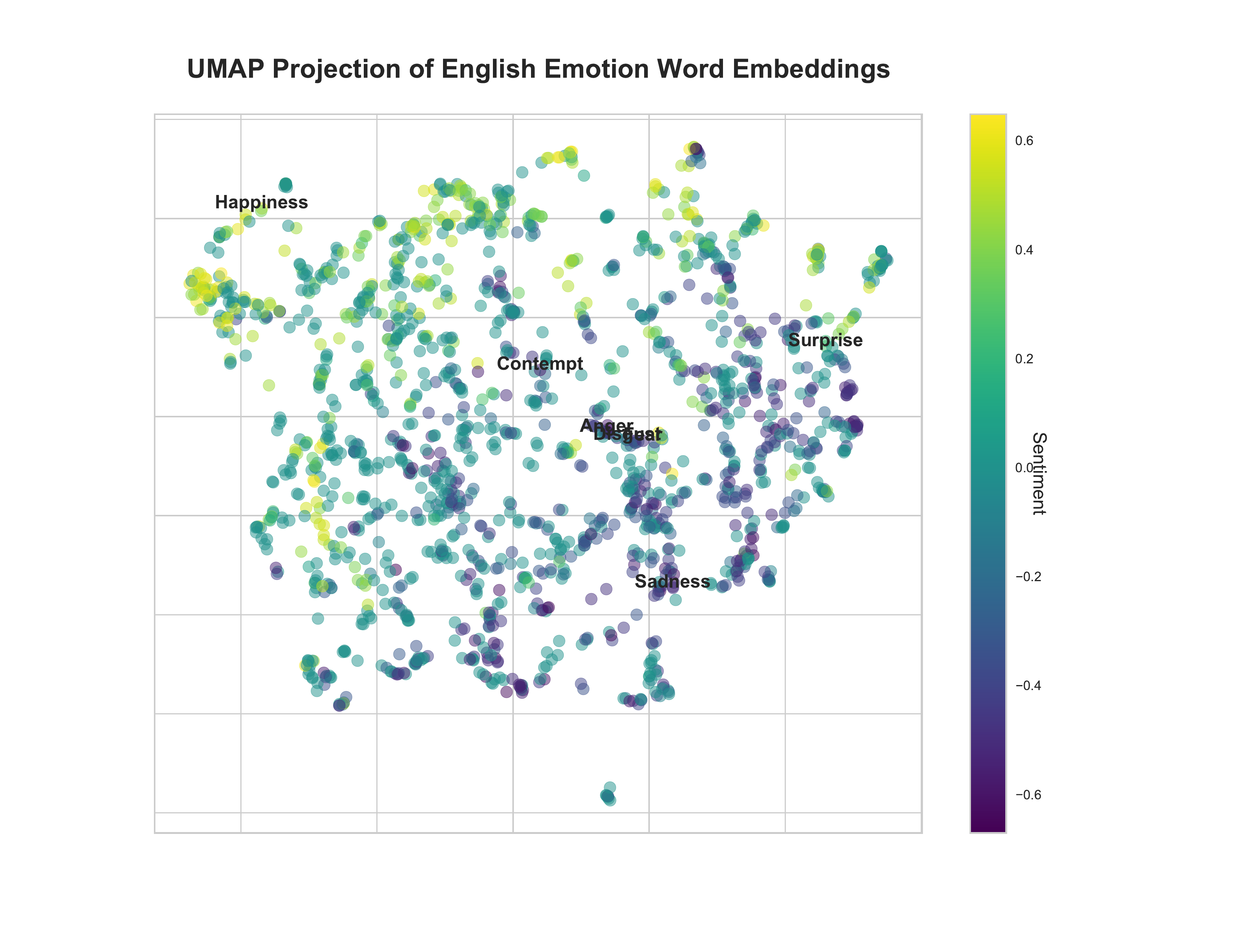}
         \end{minipage}
     \end{minipage}
    \begin{minipage}[b]{0.05\textwidth}
            \includegraphics[width=0.6\textwidth,
            trim = 0 0 0 0]{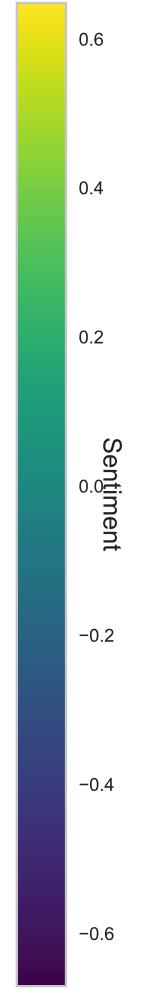}
    \end{minipage} 
    \caption{We compare four different dimensionality reduction methods using sentiment as a heuristic. Unlike t-SNE~\cite{hinton_stochastic_2003}, UMAP~\cite{mcinnes_umap_2018} is able to separate and organize positive and negative emotion concepts. Compared with other methods, when we plot Ekman's basic emotions~\cite{ekman_constants_1971} in the UMAP embedding space the separation between ``Happiness'' and the negative basic emotions is increased. This result is also shown in Table~\ref{table:umap_similarity} where we compare the cosine similarity between the raw FastText~\cite{grave_learning_2018} word vectors and UMAP embeddings across synonyms and antonyms for ``Happiness.''}
        \label{fig:umap_sentiment}
\end{figure*}

Since word embeddings are generated based on their local context, antonym word pairs ({\it e.g.}, Happiness/Sadness) which are commonly used in the same context may have high cosine similarities relative to their perceived similarity. To overcome this, we use Uniform Manifold Approximation and Projection (UMAP)~\cite{mcinnes_umap_2018}. This dimensionality reduction technique is similar to t-SNE~\cite{hinton_stochastic_2003} but has the added benefit of maintaining the global relationships among samples. Because UMAP relies on a number of neighbors to generate the embedding, as long as a word has more true synonyms than similar-context antonyms, it will close the distance between the synonyms while increasing the distance between antonyms. This outcome is demonstrated in Table~\ref{table:umap_similarity}, where UMAP increases the similarity between synonyms while pushing antonyms away. In the raw vectors, ``Sadness'' has the highest cosine similarity with ``Happiness'' despite the semantic meaning being quite different. Other dimensionality reduction techniques like principal component analysis (PCA), singular value decomposition (SVD), and t-distributed stochastic neighbor embedding (t-SNE) fail to produce accurate cosine similarities between synonyms and antonyms after reduction. 

In Fig.~\ref{fig:umap_sentiment}, we visualize the different embeddings with respect to the sentiment for each word using the Python Natural Language Toolkit (NLTK)~\cite{bird_natural_2009}. Although we don't use any sentiment information when generating the embeddings, all dimensionality reduction techniques are able to create a separation between positive and negative emotions. However, t-SNE is only able to achieve this locally and fails to provide any consistent global separation. Similar to the result in Table~\ref{table:umap_similarity}, when we plot Ekman's basic emotions, we see much clearer separation between ``Happiness'' and Ekman's negative basic emotions in the UMAP embeddings than in other techniques.

\begin{table}
\caption{UMAP's Effect on Similarity}
\label{table:umap_similarity}
\resizebox{0.48\textwidth}{!}{\begin{tabular}{lcccccc} 
\cline{3-7}
                       & \multicolumn{1}{c|}{}                  & \multicolumn{5}{c|}{\textbf{Cosine Similarity}}                                                                 \\ 
\hline
\textbf{Emotion Pairs} & \multicolumn{1}{c|}{\textbf{Synonyms}} & \begin{tabular}[c]{@{}c@{}}\textbf{Raw Word}\\\textbf{Vector}\end{tabular} & \begin{tabular}[c]{@{}c@{}}\textbf{PCA}\\\textbf{$d = 2$}\end{tabular} & \begin{tabular}[c]{@{}c@{}}\textbf{SVD}\\\textbf{$d = 2$}\end{tabular} & \begin{tabular}[c]{@{}c@{}}\textbf{t-SNE}\\\textbf{$d = 2$}\end{tabular} & \multicolumn{1}{c|}{\begin{tabular}[c]{@{}c@{}}\textbf{UMAP}\\\textbf{$d = 2$}\end{tabular}}  \\ 
\hline
Happiness, Sadness     &                                        & 0.43                                                                       & 0.38                                                                   & 0.65                                                                   & 0.88                                                                    & \textbf{-0.97}                                                                                 \\
Happiness, Anger       &                                        & 0.16                                                                       & 0.02                                                                   & 0.39                                                                   & 0.75                                                                    & \textbf{-0.92}                                                                                 \\
Happiness, Fear        &                                        & 0.13                                                                       & 0.05                                                                   & 0.39                                                                   & 0.82                                                                    & \textbf{-0.80}                                                                                 \\
Happiness, Contempt    &                                        & 0.06                                                                       & 0.61                                                                   & 0.70                                                                   & 0.82                                                                    & \textbf{-0.78}                                                                                 \\
Happiness, Disgust     &                                        & 0.10                                                                       & 0.12                                                                   & 0.56                                                                   & 0.89                                                                    & \textbf{-0.98}                                                                                 \\
Happiness, Surprise    &                                        & 0.18                                                                       & 0.98                                                                   & 0.93                                                                   & 0.89                                                                    & \textbf{-0.55}                                                                                 \\
Happiness, Joyous      & \checkmark                                      & 0.23                                                                       & -0.37                                                                  & -0.87                                                                  & 0.37                                                                    & \textbf{0.45}                                                                                  \\
Happiness, Gladness    & \checkmark                                       & 0.39                                                                       & \textbf{0.99}                                                          & 0.89                                                                   & 0.63                                                                    & \textbf{0.99}                                                                                  \\
Happiness, Bliss       & \checkmark                                       & 0.42                                                                       & 0.82                                                                   & 0.54                                                                   & \textbf{0.99}                                                           & \textbf{0.99}                                   \\   \hline \\                        
                  
\end{tabular}
}
$d$: number of dimensions
\end{table}

\subsection{Hierarchical Clustering}

\begin{figure}
    \centering
    \includegraphics[width=3.4in,trim = 0 20 0 0, clip]{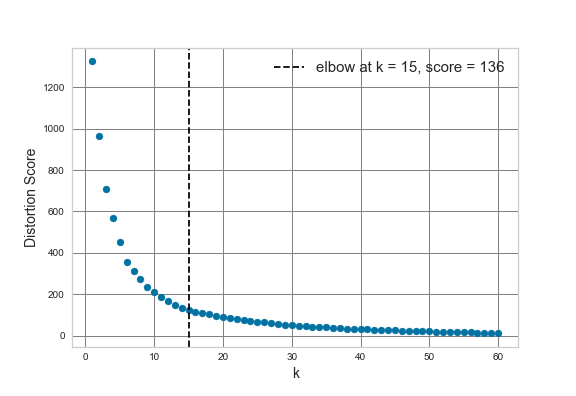}
    \caption{Using the elbow method to identify the number of clusters for each culture for the cross-cultural model of emotion. $k$ is the number of clusters.}
    \label{elbow}
\end{figure}

\begin{figure*}[ht!]
 \centering
\includegraphics[width=0.95\textwidth, trim=6cm 4cm 6cm 0cm, clip]{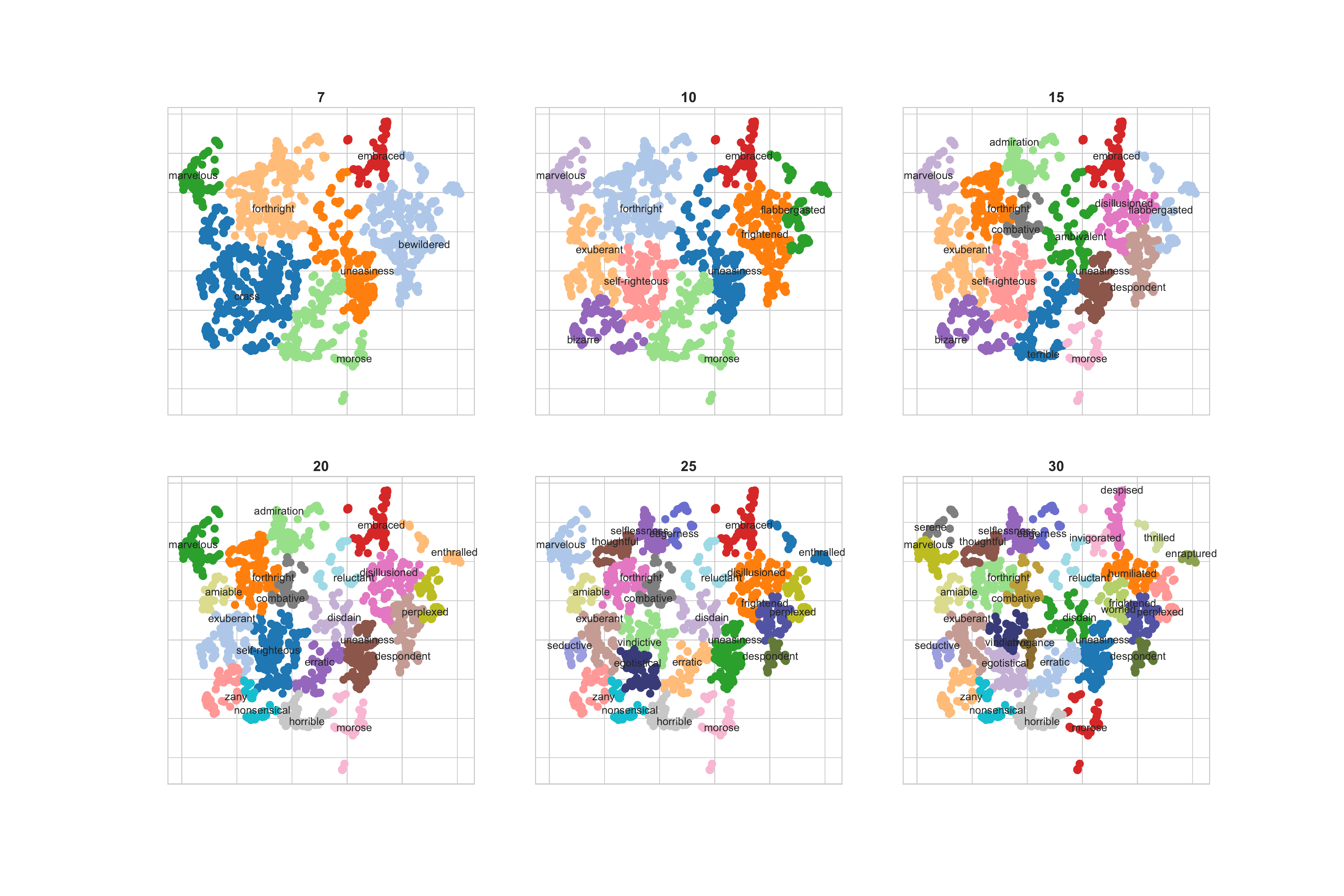}
 \caption{Cluster summaries for English for the number of clusters, $k$, increases from seven to 30. As $k$ increases, we are able to generate more complete models of human emotion. Since UMAP maintains global relationships among samples, we can visualize the relationships among different emotions.
 }
 \label{fig:en_clusters}
\end{figure*}

Agglomerative clustering~\cite{maimon_data_2005, dalmia_clustering_2021} was used to programmatically select emotion words with the highest coverage. We chose agglomerative clustering over other methods because it mirrors the hierarchical nature of basic emotions~\cite{ekman_argument_1992, panksepp_affective_2004}. To do this, we first determine the optimum number of clusters using the elbow method~\cite{thorndike_who_1953}. This automatically finds the ``elbow'' or ``knee'' which corresponds to point of maximum curvature on the plot of cluster distortion by clusters count~\cite{satopaa_finding_2011} as shown in Fig.~\ref{elbow}. Finally, we summarize the contents of each group by finding the word embedding closest to the centroid of each cluster. The results of this step are shown in Fig.~\ref{fig:en_clusters}.

To develop our cross-cultural model, we repeat this process by translating our list of emotion-related concepts from English into the other official languages recognized by the United Nations (Arabic, Chinese, French, Spanish, and Russian) using Google's Translation API.\footnote{https://cloud.google.com/translate} We then once again proceeded with Facebook's FastText models which have been trained on Common Crawl and Wikipedia for each of these languages~\cite{grave_learning_2018}.  

\begin{table}
\centering
\caption{Recommended Number of Clusters from the Elbow Method}
\label{table:elbow}
\begin{tabular}{lc} 
\hline
\textbf{Language} & \begin{tabular}[c]{@{}c@{}}\textbf{\# Clusters}\\\textbf{(Elbow)}\end{tabular}  \\ 
\hline
English           & 14                                                                                       \\
Arabic            & 14                                                                                       \\
Mandarin Chinese  & 15                                                                                       \\
French            & 11                                                                                       \\
Spanish           & 13                                                                                       \\
Russia            & 14                                                                                       \\ \hline
Cross-Cultural    & 15       \\
\hline

\end{tabular}
\end{table}

The translated UMAP embeddings for both Chinese and Russian needed adjustments due to limitations in translation. For Chinese, the embeddings originally formed two distinct clusters. This separation is caused by the inclusion of the nominalization particle ``\begin{CJK}{UTF8}{gbsn}的\end{CJK},'' which
converts nouns or noun phrases into adjectives ({\it e.g.}, ``\begin{CJK}{UTF8}{gbsn}快乐\end{CJK}'' or happiness $\rightarrow$ ``\begin{CJK}{UTF8}{gbsn}快乐的\end{CJK}'' or happy). 
An equivalent example in English would be the use of the suffix ``-ness'' such as in ``happiness.'' This suffix converts the adjective ``happy'' into a noun. Since FastText uses subword information to generate its word vectors, the semantic meaning of this character influences the final word embedding. When UMAP is run on these embeddings, the inclusion of this character creates enough separation between the word vectors that all words containing this character get clustered separately from the rest of the emotion-concepts list. To resolve this divide, this cluster was excluded from the analysis since the other was significantly larger and already contained a superset of Ekman's basic emotions. In Russian, there was a similar separation caused by a noun-adjective divide. Here, the noun cluster was excluded because it was roughly a quarter the size of the adjective cluster. 

With the translated word vectors, we once again apply agglomerative clustering to each language. Since this process combines the two closest centroids in each iteration, this clustering can handle situations where one-to-one translations do not exist. As shown in Table~\ref{table:elbow}, we start experiencing diminishing returns in all languages around 15 clusters. This result provides quantitative evidence that all other emotion-related words can be embedded into the semantic space of roughly 15 dimensions.

\begin{table}[ht!]
\centering
\caption{HICEM Components in Six Major Languages}
\label{table:components}
\includegraphics[width=0.48\textwidth, trim=0 0 0 4mm, clip]{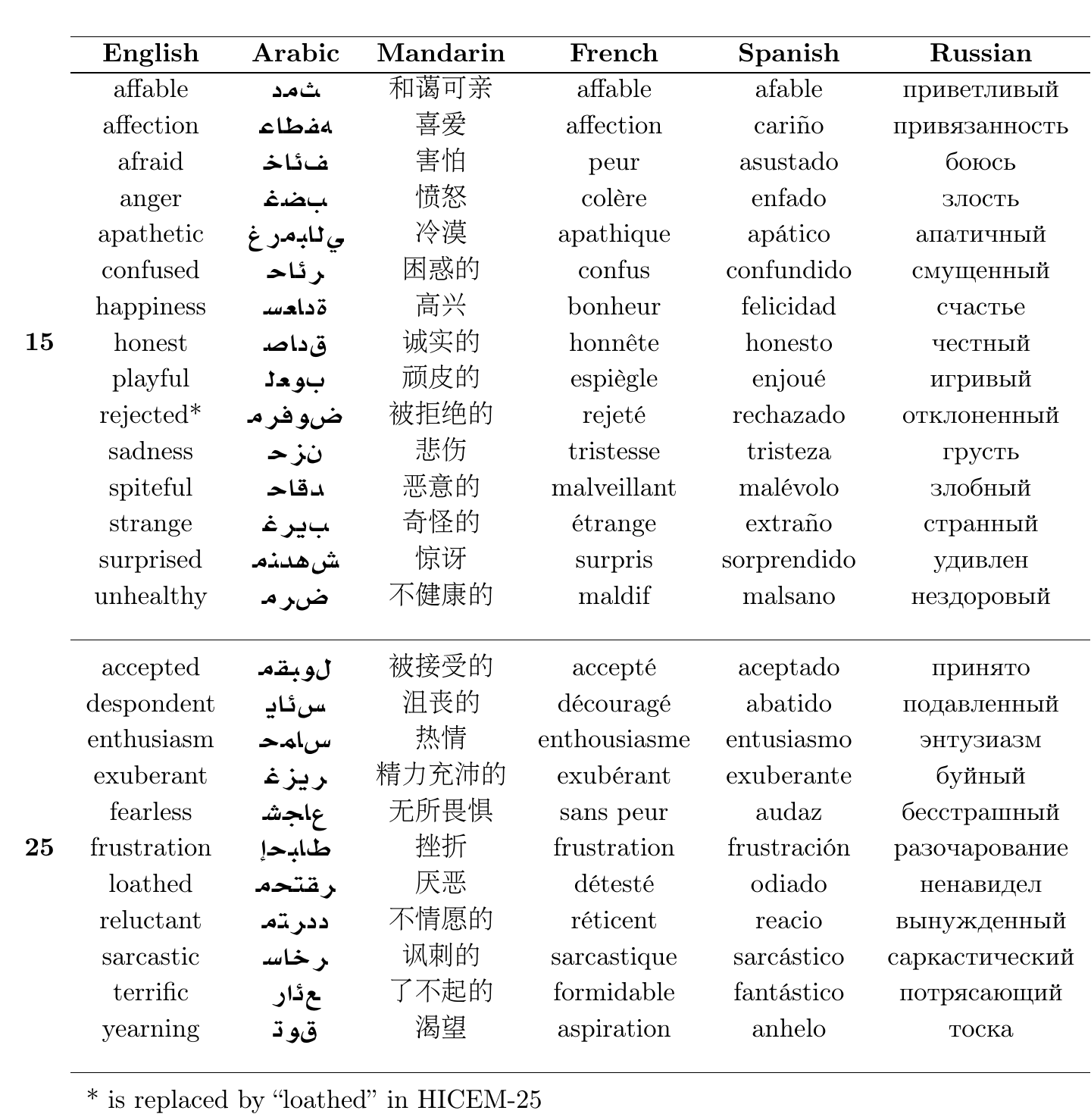}

\end{table}

An interesting finding from this analysis is that these clusters, broadly capture the basic emotions as described by Ekman ({\it i.e.}, fear,
anger, happiness/joy, sadness, disgust, and surprise) in each language. Likewise, there seem to be themes of humor, abnormal behavior, friendliness, anxiety, and confusion. To ensure our model has coverage across cultures, we then took the top 50 cluster summary words from each language and once again performed agglomerative clustering ($k$=15) to generate a list of cross-cultural summary words. As a final step, we then manually adjusted our emotion model by replacing rare or outdated language with more common English terms. This process provides us with the final list of 15 high coverage emotion concepts in Table~\ref{table:components}. We repeat this final agglomerative clustering step for 25 clusters (named HICEM-25) to provide an additional comparison with similar-sized emotion models. 

\subsection{Assessing Performance}

\begin{table*}
\caption{Coverage across Languages and Emotion Models}
\label{table:results}
\resizebox{\textwidth}{!}{\begin{tabular}{lcccccccc} 
\cline{3-9}
                       & \multicolumn{1}{l|}{}                         & \multicolumn{7}{c|}{\textbf{Average Coverage}}                                                                                                                                    \\ 
\hline
\textbf{Emotion Model} & \multicolumn{1}{c|}{\textbf{\# Components}} & \textbf{English}     & \textbf{Arabic}      & \textbf{Mandarin}    & \textbf{French}      & \textbf{Spanish}     & \textbf{Russian}     & \multicolumn{1}{c|}{\textbf{Total}}  \\ 
\hline
Ekman~                 & 7                                             & 0.324                & 0.273                & 0.286                & 0.355                & 0.333                & 0.327                & 0.314                                \\
EMOTIC~                & 27*                                           & 0.400                & 0.374                & 0.332                & 0.409                & 0.397                & 0.437                & 0.390                                \\
Cowen~                 & 27                                            & 0.432                & 0.352                & 0.332                & 0.443                & 0.447                & 0.505                & 0.415                                \\
GoEmotions~            & 28\textsuperscript{†}                         & 0.430                & 0.362                & 0.332                & 0.449                & 0.448                & 0.479                & 0.414                                \\
Plutchik~              & 32                                            & 0.461                & 0.365                & 0.341                & 0.457                & 0.475                & 0.503                & 0.428                                \\ 
\hline
\textcolor{navyblue}{HICEM-15}               & 15                                            & 0.458                & 0.349                & 0.315                & 0.455                & 0.458                & 0.501                & 0.416                                \\
\textcolor{navyblue}{HICEM-25}              & 25                                            & 0.503                & 0.378                & 0.337                & 0.484                & 0.490                & 0.528                & 0.444                                \\ \hline \\
                   
\end{tabular}}
*The combined category Doubt/Confusion was split into two labels ~~\textsuperscript{†}27 Emotion labels + Neutral
\end{table*}

We can qualitatively assess the quality of this list, by projecting it into the English embeddings as shown in Fig.~\ref{fig:model comparison}. Because this list is uniformly distributed throughout the space, we can assume there is minimal overlap in its labels. In addition to this, we also define a new metric ``coverage'' to provide a quantitative assessment of the quality of each model. This metric is based on cosine similarity between the words in the model and our emotion list as follows
$$\text{Avg. Coverage}(M) = \frac{1}{n}\sum_{i=1}^{n} \max_{m\epsilon M} \text{CosSim}(m, w_i)\;,$$
and 
$$ \text{CosSim}(a, b) = \frac{a \cdot b}{\lVert a \rVert \lVert b \rVert}\;.$$ 
\begin{figure}[ht!]

    \includegraphics[width=0.5\textwidth, trim = 35 10 35 20, clip]{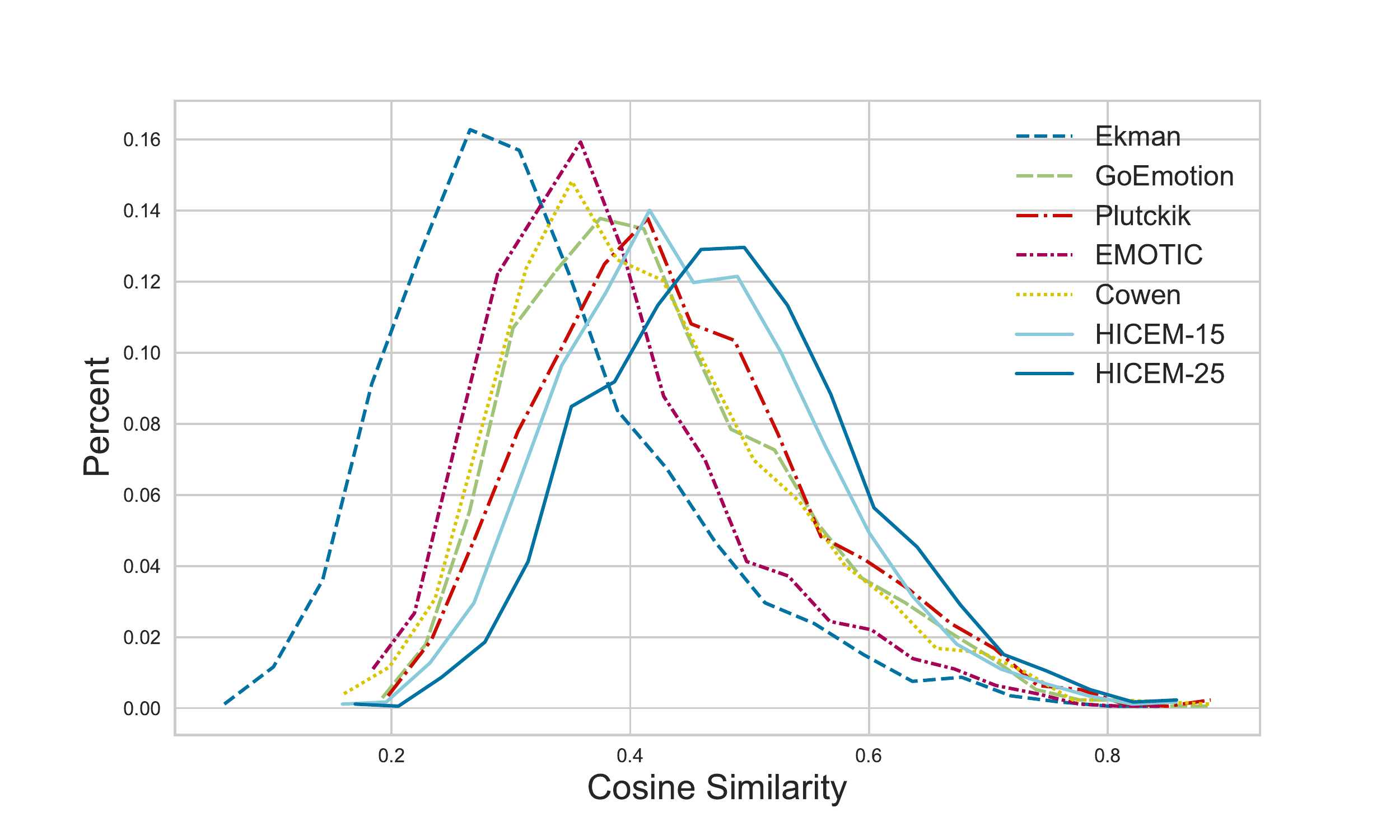}
    
    \caption{In this histogram of coverage by cosine similarity across English word vectors, we show that our HICEM-15 model is able to provide higher coverage with fewer labels compared to other models of emotion popular in psychology~\cite{ekman_constants_1971,kort_affective_2001, plutchik_emotions_1991, kosti_emotic_2017, cowen_self-report_2017, demszky_goemotions_2020}. In English, our model achieves an average similarity of 0.45. This result is equivalent to the similarity between ``happiness'' and ``calmness,'' ``merriment,'' or ``euphoria'' ($0.45 \pm 0.01$).
    }\label{fig:results}

\end{figure}

Here $M$ is the set of FastText word vectors in an emotion model, and  $W-M $ is the set of vectors for the words from the set of 1,720 emotion concepts (the set $W$) excluding the components of the emotion model (the set $M$). We define $n$ as the cardinality of set $W - M$. We chose to proceed with max cosine similarity since this gives an indication of how strong the top synonym is for a given emotion model and emotion-concept pair. As seen in Table~\ref{table:results} and Fig.~\ref{fig:results}, our model provides higher coverage with fewer components compared with previous models.

In addition to analyzing the coverage, we also conducted an experiment to show how much emotion information is captured in each model. The intuition behind this experiment is that ideally, we should be able to recover a large number of emotional states for a sample using only the annotations from a given model. In contrast with coverage where we only consider the max cosine similarity, here we leverage the additional annotations across all the components to make a prediction on the emotion present. To do this, we assume we have an oracle who, when given an emotion model $M$, is able to generate an embedding $X$ for a given word vector $w_i$ using the cosine similarities between the FastText word vectors of each $m \in M$ and $w_i$. More formally,
$$X = \text{CosSim}(w_i, M)\;. $$
Then, 
$$\hat{w_i} = G(X)\;,$$ 
where $G$ is a function we pass $X$ through in order to make a prediction on the original word vectors $\hat{w_i}$. Then once again using cosine similarity, we compare the original word vector $w_i$ with our recovered prediction $\hat{w_i}$ as follows:
$$ \text{Avg. Recovered Information}(M) = \frac{1}{n}\sum_{i=1}^{n} \text{CosSim}(w_i, \hat{w_i})\;.$$

In practice, if we are to annotate the emotional expression of a character in a video clip using a categorical emotion model, this would be equivalent to a human annotator for the clip judging how similar or dissimilar the sample ({\it i.e.}, the character's emotional expression) is to the labels of the given emotion model and then using those annotations to recover the ground-truth emotion present in the clip. There is often a level of disagreement among annotators in real-world settings which adds noise to the oracle's embedding $X$. However, the assumption of a perfect oracle provides a theoretical upper bound for the amount of recoverable information which we can use for comparison.

\begin{figure}[ht!]
    \includegraphics[width=0.5\textwidth,
        trim = 30 10 30 30, clip]{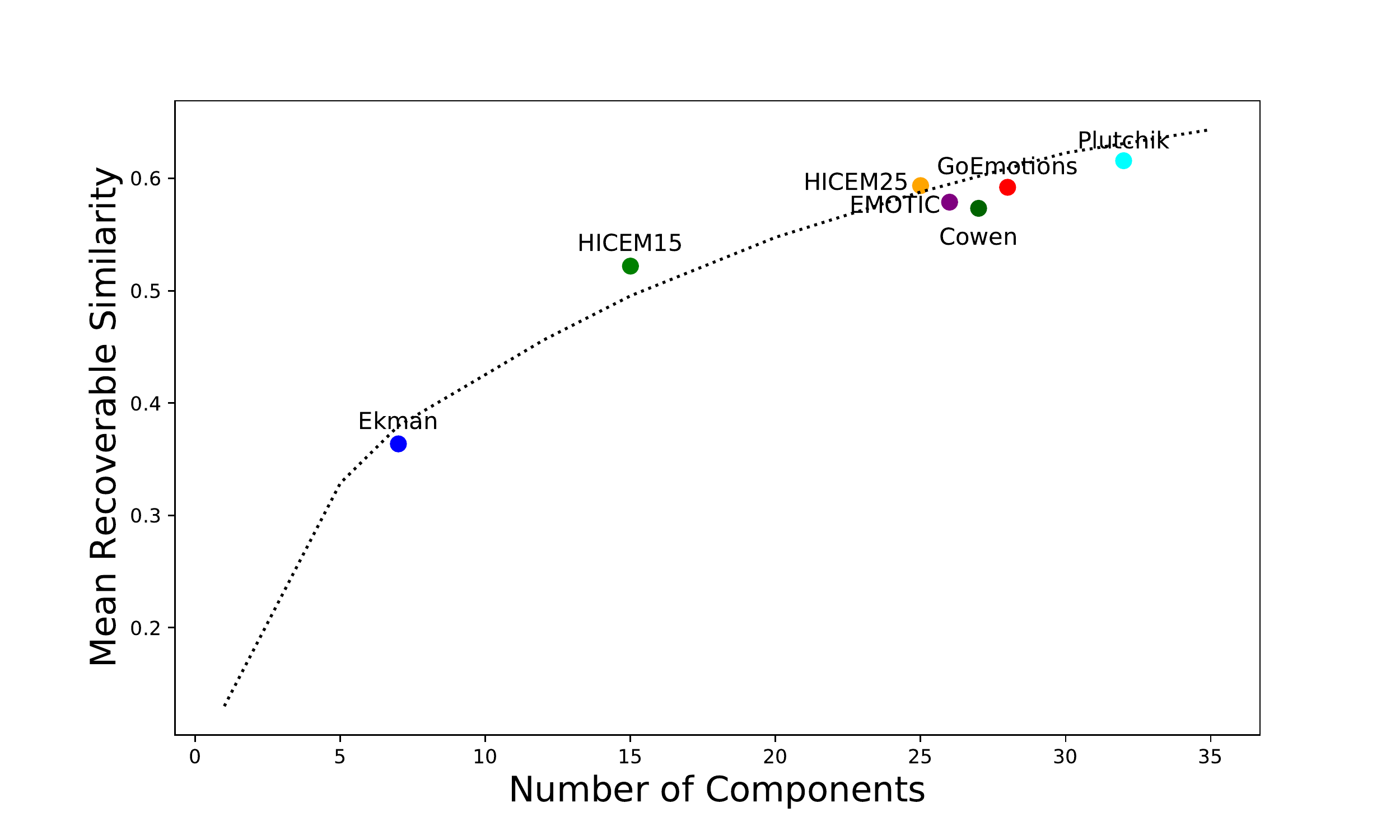}
    \caption{In this plot of recovered cosine similarity vs. the number of components using English word vectors, the only models which outperform a uniform random sample of words ({\it i.e.}, the dotted line) are HICEM-15 and HICEM-25.}
    
    \label{fig:recovered}
\end{figure}

In our experiment, we use the Ridge regression~\cite{hoerl_ridge_1970} for $G$ and a 50/50 train-test split on the list of 1,720 emotion concepts. We then train on the similarity embeddings for each model to make a prediction on the original embedding. To provide a baseline, we use random subsets of words from the emotion-concept list of varying sizes. As shown in Fig.~\ref{fig:recovered} and Table~\ref{table:resultsRecovered}, HICEM is the only emotion model that performs above a random subset of the same size. There are several possible explanations for this. First, this model's success could be influenced by the presence of redundant or overlapping labels. Second, because the random subsets were selected using a random uniform distribution, they are more likely to sample across the entire emotion concept space, in turn providing more unique information. Lastly, HICEM benefits from being biased towards the emotion-concept list it is being tested against. Although other models touch on concepts such as craving~\cite{cowen_self-report_2017} and pain~\cite{luo_arbee_2020, kosti_emotic_2017}, these types of concepts have only recently been included in the discussion of emotion~\cite{dancy_hybrid_2021, al-shawaf_evolutionary_2016, gilam_what_2020}. Since these were included in the list of 1,720 emotion concepts, HICEM is better able to represent them since it is derived from this same list. 

\begin{figure*}[ht!]
    \centering
    \includegraphics[width=0.95\textwidth, trim = 4cm 4cm 4cm 3cm, clip]{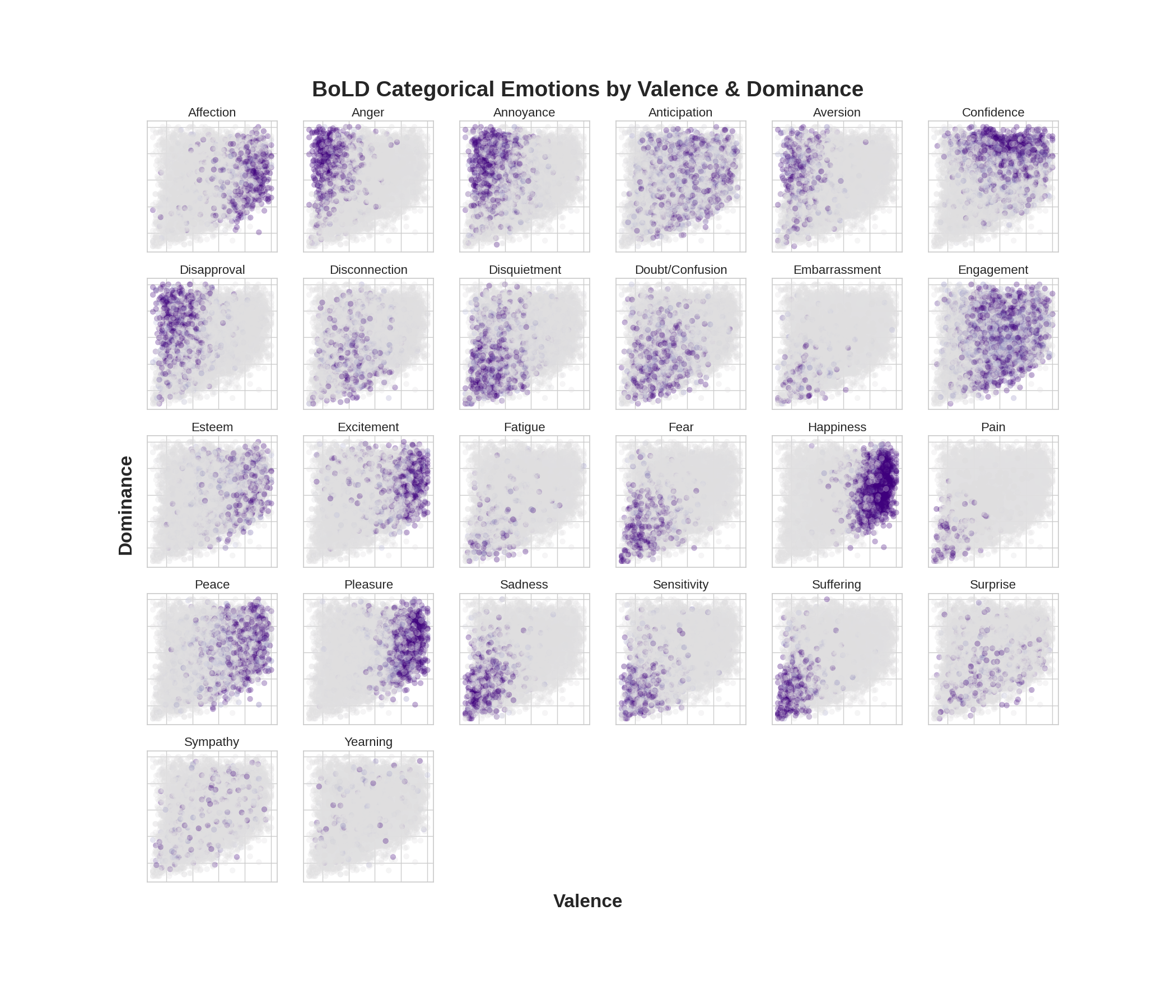}
    \caption{Categorical emotions plotted against valence and dominance for the BoLD dataset. Horizontal delineation suggests valence and vertical delineation suggests dominance as the discriminant factor. In these scatterplots of categorical emotions according to their valence and dominance ratings, there appears to be a gap ({\it i.e.}, blank space) for low-dominance, high-valence emotions. Looking at the few samples present in this area; these seem to represent guardian-child type relationships where one subject takes on the role of the protector and the other takes on the role of the child. We suggest the lack of samples in this area is due to some bias in the original data collection and not necessarily due to a limitation of the continuous VAD dimensions.}
    \label{s:bold_ValDom_scatter}
\end{figure*}

It is worth noting that in practice, there is a trade-off between the size of the emotion model and the amount of information the annotators will give for any sample. As emotion models become more complex and include more abstract concepts, the agreement between annotators decreases substantially. This reduction can be seen in the levels of inter-annotator agreement across several large-scale datasets that reported this information. For example, despite several quality-control measures implemented during data collection and the post-processing done to filter unreliable annotators in the BoLD dataset~\cite{luo_arbee_2020}, the average Fleiss' Kappa~\cite{gwet_handbook_2014} across emotion categories is $\kappa=0.173$~\cite{luo_arbee_2020}. Intuitively less complex emotions like ``Happiness'' have higher levels of agreement comparable to objective tasks performed at the time of data collection like determining age or ethnicity. More abstract concepts, such as ``Yearning'' and ``Sensitivity,'' had almost no agreement among participants. This result mirrors those from the EMOTIC~\cite{kosti_emotic_2017} and GoEmotions~\cite{demszky_goemotions_2020} datasets. This comparison is important since as the size of the emotion model increases the types of emotion concepts included are bound to become more abstract. Not only do additional components have diminishing returns in terms of the information they provide, but since emotion is subjective, they also suffer an agreement penalty during the annotation process further reducing their effectiveness. One method to help mitigate this effect would be to choose more concrete emotion concepts as the foundation for emotion models. This method wasn't considered when generating the random models in Table~\ref{table:resultsRecovered}. Although they seem to outperform existing models in terms of recoverable information, the abstractness of their labels would severely limit their real-world effectiveness.

\begin{table*}
\caption{Recoverable Information across Languages and Emotion Models}
\label{table:resultsRecovered}
\resizebox{\textwidth}{!}{\begin{tabular}{lcccccccc} 
\cline{3-9}
                       & \multicolumn{1}{c|}{}                         & \multicolumn{7}{c|}{\textbf{Average Recoverable Information}}                                                                                                                      \\ 
\hline
\textbf{Emotion Model} & \multicolumn{1}{c|}{\textbf{\# Components}} & \textbf{English}     & \textbf{Arabic}      & \textbf{Mandarin}    & \textbf{French}      & \textbf{Spanish}     & \textbf{Russian}     & \multicolumn{1}{c|}{\textbf{Total}}  \\ 
\hline
Ekman~                 & 7                                             & 0.370                & 0.263                & 0.336                & 0.328                & 0.325                & 0.339                & 0.327                                \\
EMOTIC                 & 27*                                           & 0.585                & 0.470                & 0.502                & 0.515                & 0.534                & 0.528                & 0.522                                \\
Cowen                  & 27                                            & 0.577                & 0.414                & 0.497                & 0.512                & 0.545                & 0.529                & 0.512                                \\
GoEmotions             & 28\textsuperscript{†}                         & 0.592                & 0.473                & 0.504                & 0.540                & 0.544                & 0.552                & 0.534                                \\
Plutchik~              & 32                                            & 0.620                & 0.485                & 0.525                & 0.552                & 0.573                & 0.562                & 0.552                                \\ 
\hline
Random-7               & 7                                             & 0.372                & 0.312                & 0.346                & 0.332                & 0.340                & 0.386                & 0.348                                \\
Random-15              & 15                                            & 0.507                & 0.393                & 0.435                & 0.442                & 0.468                & 0.497                & 0.457                                \\
Random-25              & 25                                            & 0.597                & 0.468                & 0.477                & 0.522                & 0.544                & 0.550                & 0.526                                \\
Random-30              & 30                                            & 0.625                & 0.492                & 0.491                & 0.550                & 0.566                & 0.564                & 0.548                                \\ 
\hline
\textcolor{navyblue}{HICEM-15}                & 15                                            & 0.520                & 0.412                & 0.426                & 0.464                & 0.473                & 0.494                & 0.464                                \\
\textcolor{navyblue}{HICEM-25}                & 25                                            & 0.591                & 0.451                & 0.482                & 0.537                & 0.548                & 0.556                & 0.528                                \\ \hline \\
                               
\end{tabular}}
*The combined category Doubt/Confusion was split into two labels ~~\textsuperscript{†}27 Emotion labels + Neutral
\end{table*}

\section{Large-Scale Data Analysis}\label{section:largeScale}

To gain additional insight into how people consciously perceive emotion, we also examined the annotations for two large-scale emotion recognition datasets: EMOTIC~\cite{kosti_emotic_2017} and BoLD~\cite{luo_arbee_2020}. These are both in-the-wild datasets annotated for the same 26 categorical emotions as well as valence, arousal, and dominance~\cite{mehrabian_pleasure-arousal-dominance_1996}. Although the categorical emotions were not based on any pre-existing emotion models, there is enough overlap with these emotion models to assess their validity. In addition to this, since EMOTIC and BoLD are image and video data, respectively, together they offer some insight into how humans process these different modalities.

\begin{figure*}[ht!]
    \centering
    \vspace{0.5cm}
    \textbf{Projections of BoLD Categorical Labels Across Continuous Dimensions}
    
    \begin{minipage}{0.33\textwidth}
        \centering
        \vspace{3mm}
        Circumplex Projection
        \includegraphics[width=\textwidth,
        trim = 50 50 50 60, clip]{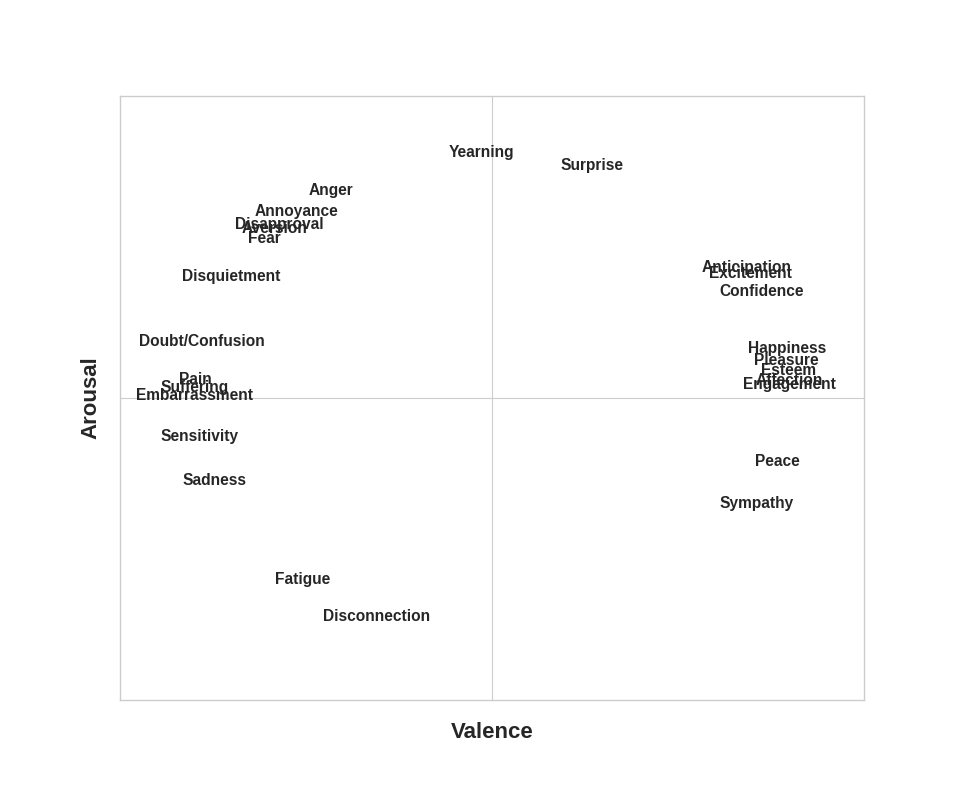}
    \end{minipage}
    \hfill
    \begin{minipage}{0.33\textwidth}
        \centering
        \vspace{3mm}
        Raw Valence-Arousal Projection
        \includegraphics[width=\textwidth,
        trim = 50 50 50 60, clip]{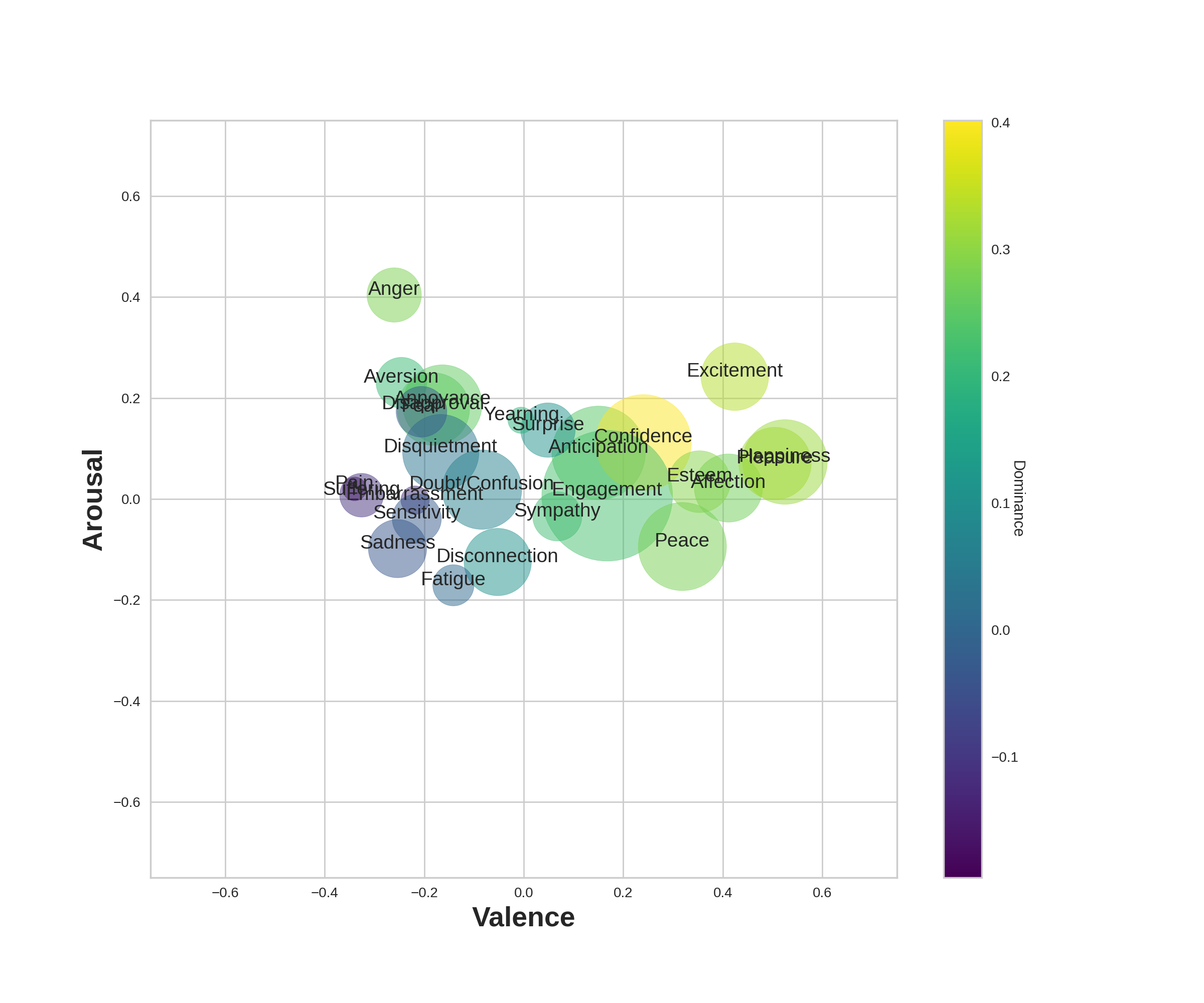}
    \end{minipage}
    \hfill
    \begin{minipage}{0.33\textwidth}
        \centering
        \vspace{3mm}
        Raw Dominance-Valence Projection
        \includegraphics[width=\textwidth,
        trim = 50 50 50 60, clip]{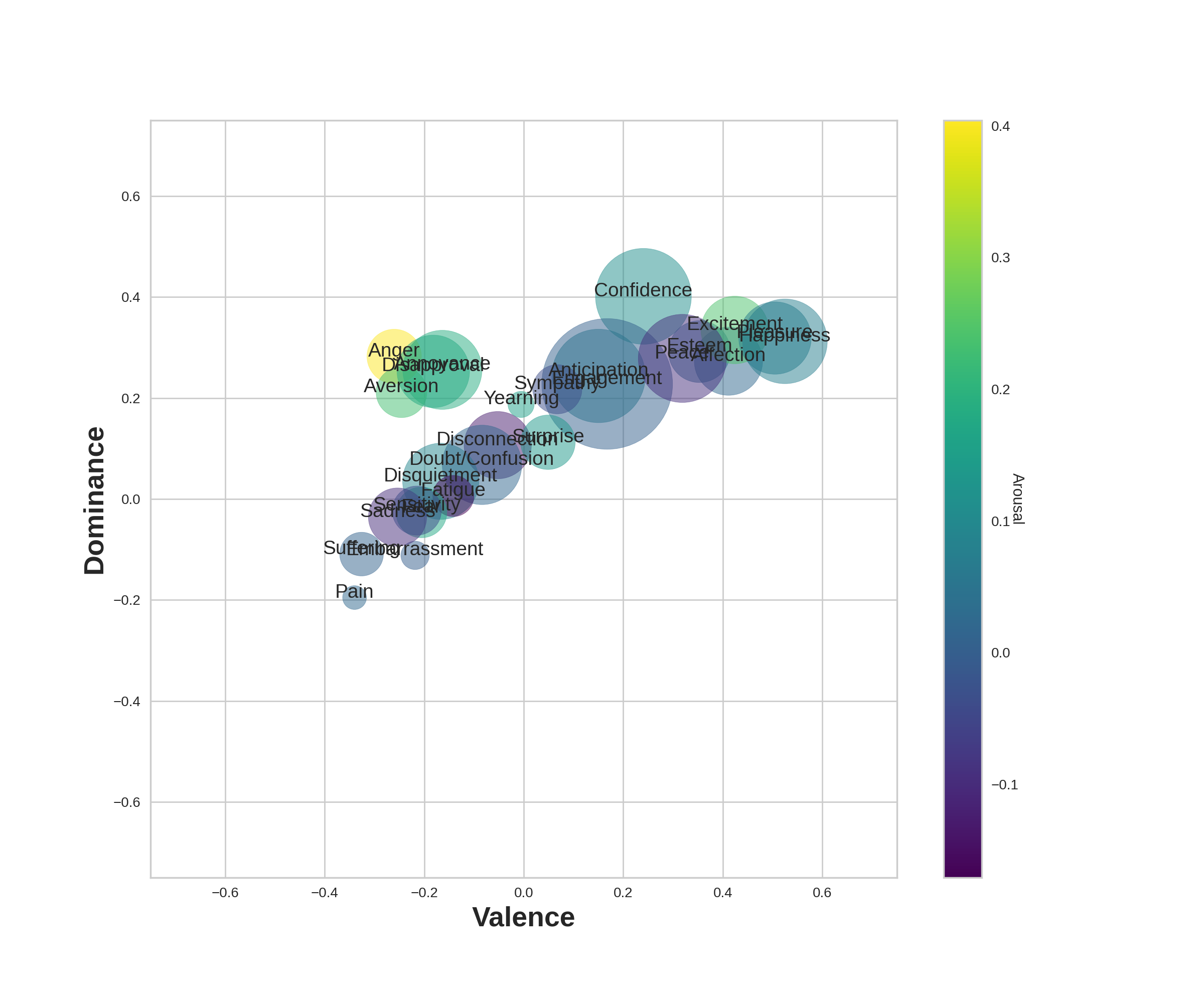}
    \end{minipage}
    \caption{{\bf Left:} When the BoLD labels are projected onto the unit circle for valence and arousal, the locations of the emotions broadly line up with the Affective Circumplex~\cite{russell_circumplex_1980}. {\bf Center:} The relationship between raw Valence-Arousal dimensions is visualized. These are colored by Dominance. The circle size is based on the number of samples for that label within the BoLD dataset. The compression of emotions along the Arousal dimension is in line with previous experiments~\cite{feldman_valence_1995}. {\bf Right:} Similar to previous work~\cite{warriner_norms_2013}, the Dominance-Valence projection shows a strong linear relationship between these two dimensions. }
    \label{fig:circumplex}
\end{figure*}

\begin{figure*}
 \centering
 \textbf{BoLD UMAP Embeddings in the Categorical Space\vspace{0.3cm}}
 
 \includegraphics[width=\textwidth,
        trim = 3.4cm 4.5cm 2cm 4.5cm, clip]{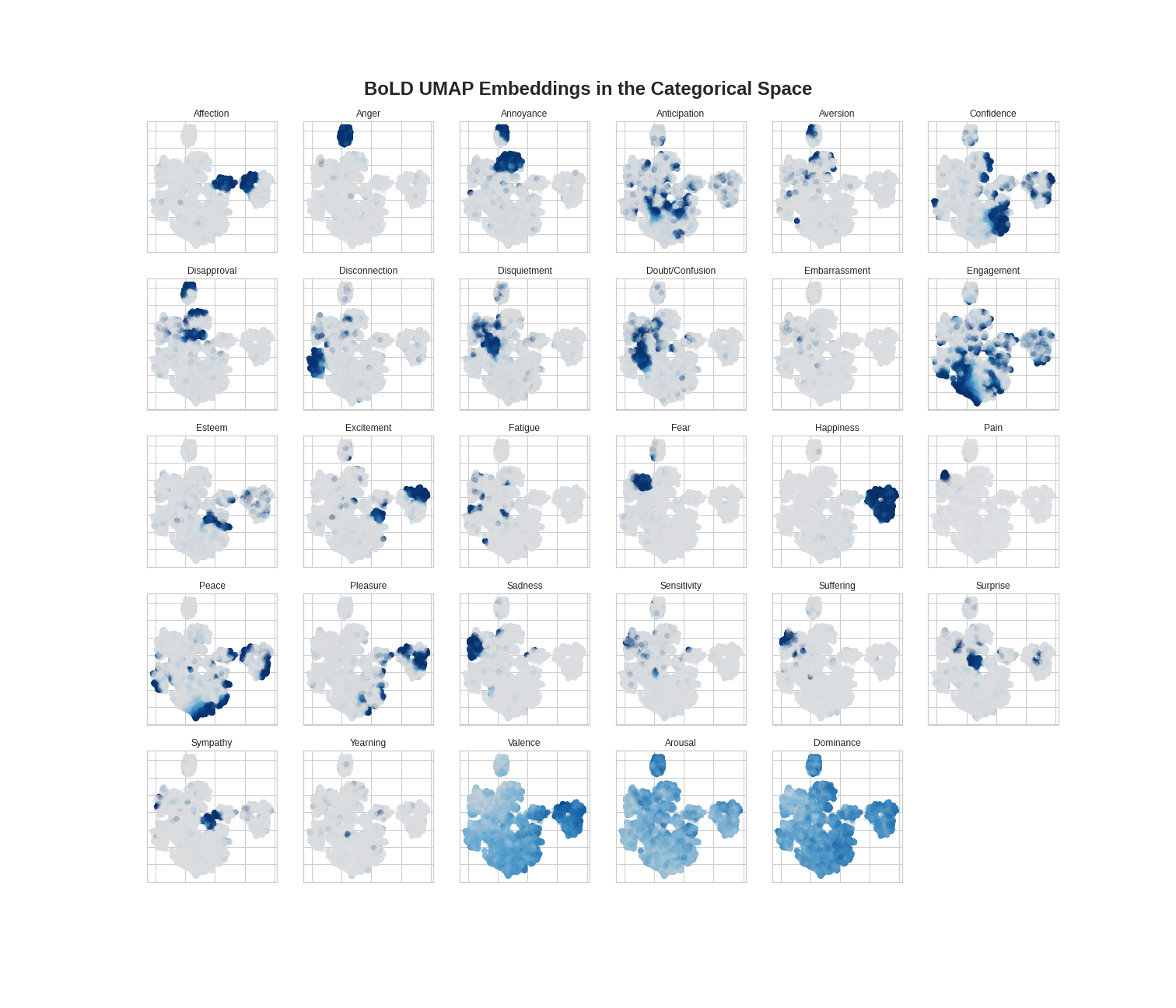}
 \caption{Heatmaps for each of the 26 categorical emotions in the BoLD dataset based on their categorical UMAP embeddings. The blue areas represent where that emotion exists within this embedding. Ekman's basic emotions form concrete clusters suggesting they are the building blocks for more complex emotions such as Annoyance or Pleasure. Similarly, other higher-order emotions such as Anticipation and Engagement appear to spread out throughout the embeddings representing the wide range of scenarios in which they present themselves. Although the axes are meaningless, by projecting the continuous VAD dimension in this space (last three images) we can roughly correlate Valence and Dominance to the X-axis and Arousal to the Y-axis.}
 \label{fig:bold_umap}
\end{figure*}

\begin{figure}
    \centering
    \includegraphics[width=0.5\textwidth, trim = 4cm 4cm 4cm 3cm, clip]{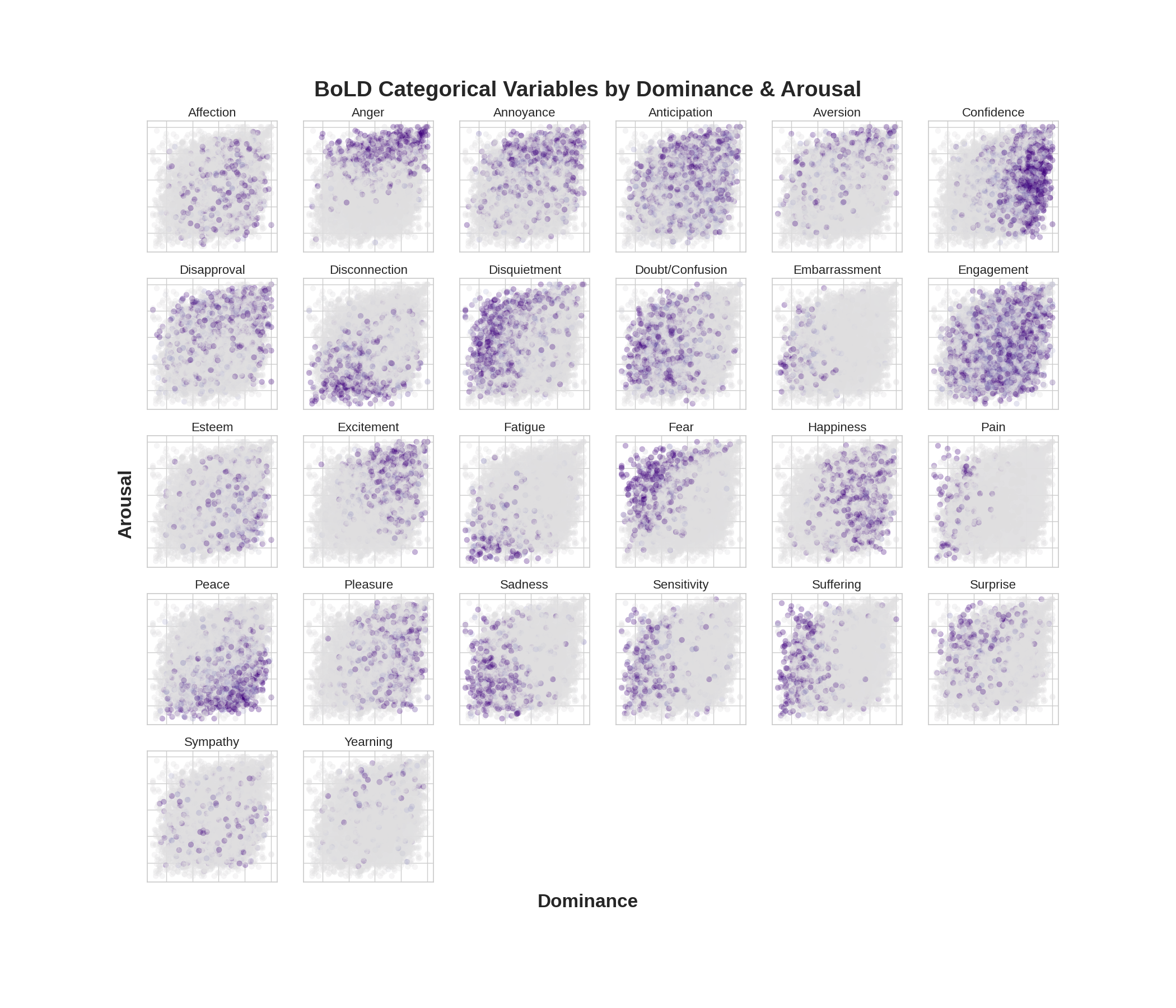}
    \caption{Categorical emotions plotted against dominance and arousal for the BoLD dataset. Horizontal delineation suggests dominance and vertical delineation suggests arousal as the discriminant factor.}
    \label{s:bold_DomAr_scatter}
\end{figure}

\begin{figure}
    \centering
    \includegraphics[width=0.5\textwidth, trim = 4cm 4cm 4cm 3cm, clip]{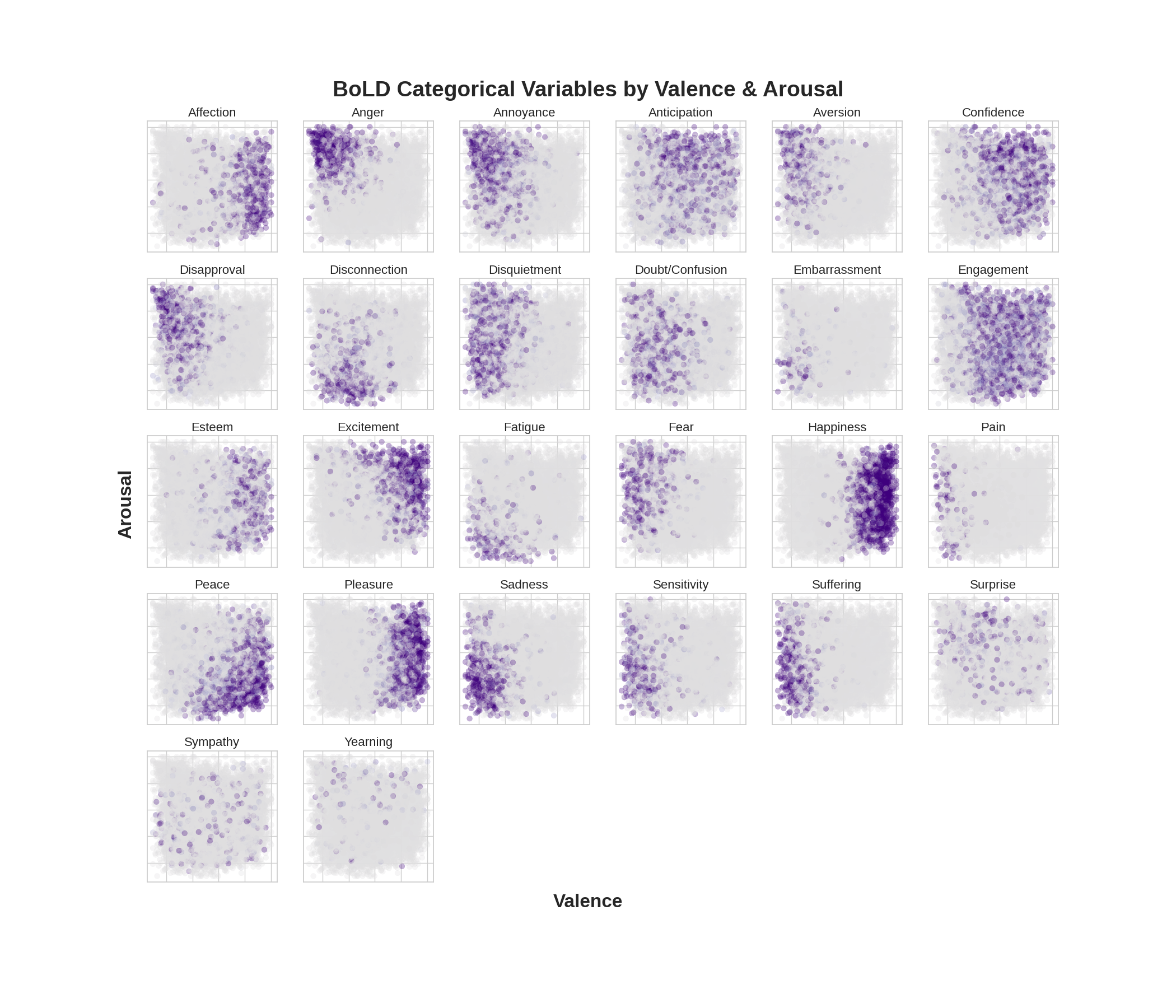}
    \caption{Categorical emotions plotted against valence and arousal for the BoLD dataset. Horizontal delineation suggests valence and vertical delineation suggests arousal as the discriminant factor.}
    \label{s:bold_ValAr_scatter}
\end{figure}

\begin{figure}
    \centering
    \includegraphics[width=0.5\textwidth, trim = 4cm 4cm 4cm 3cm, clip]{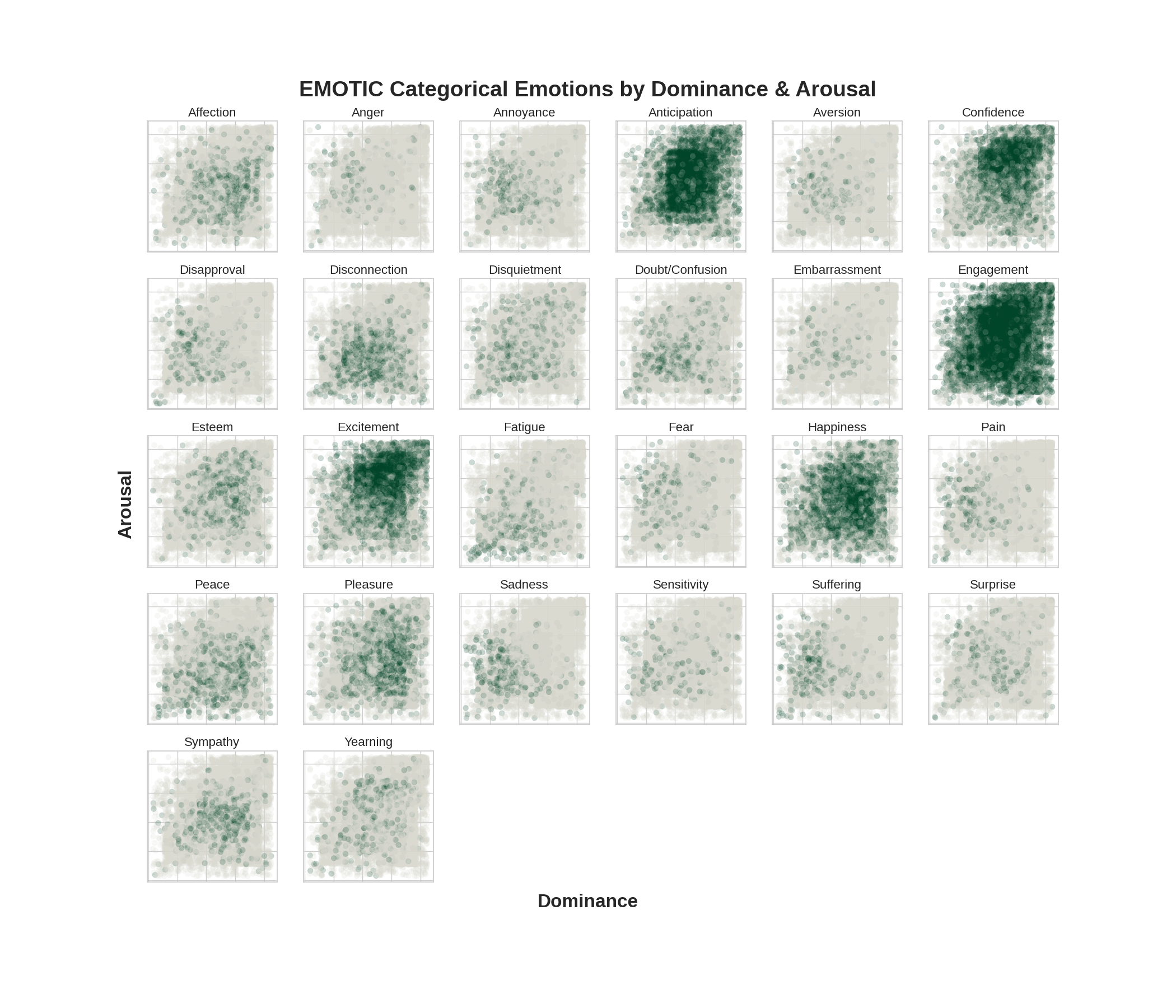}
    \caption{Categorical emotions plotted against dominance and arousal for the EMOTIC dataset. Horizontal delineation suggests dominance and vertical delineation suggests arousal as the discriminant factor.}
    \label{s:emotic_DomAr_scatter}
\end{figure}

\begin{figure}
    \centering
    \includegraphics[width=0.5\textwidth, trim = 4cm 4cm 4cm 3cm, clip]{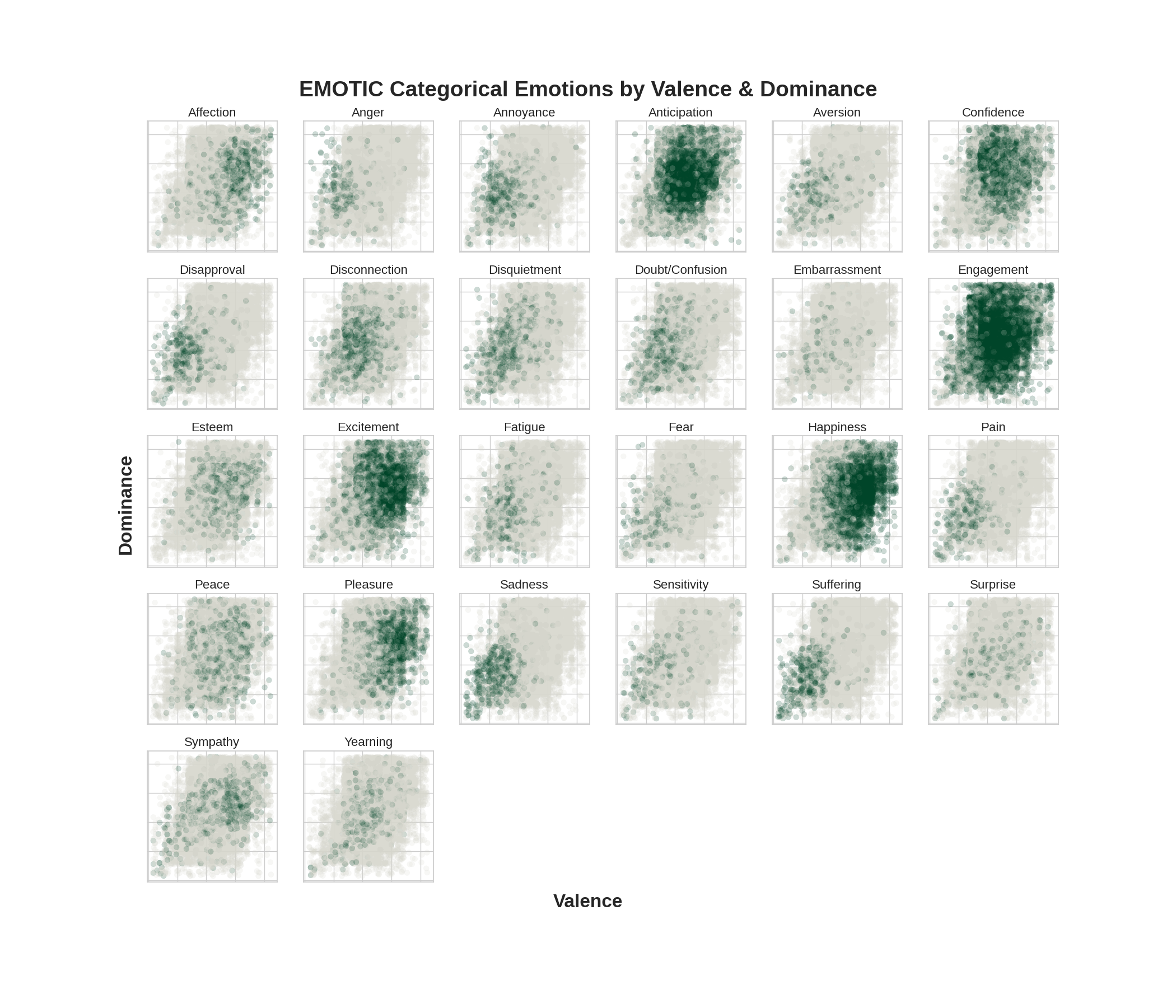}
    \caption{Categorical emotions plotted against valence and dominance for the EMOTIC dataset. Horizontal delineation suggests valence and vertical delineation suggests dominance as the discriminant factor.}
    \label{s:emotic_ValDom_scatter}
\end{figure}

\begin{figure}
    \centering
    \includegraphics[width=0.5\textwidth, trim = 4cm 4cm 4cm 3cm, clip]{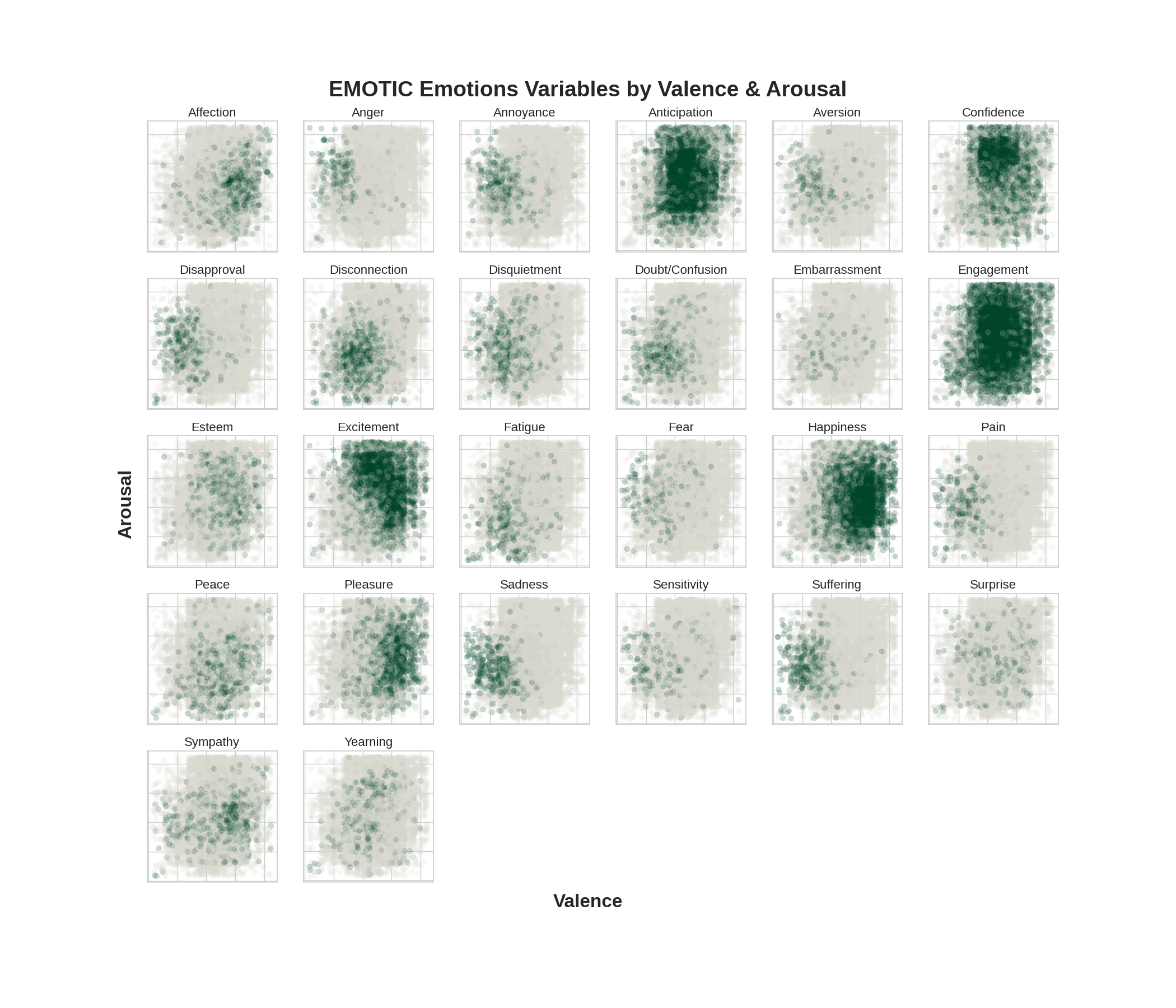}
    \caption{Categorical emotions plotted against valence and arousal for the EMOTIC dataset. Horizontal delineation suggests valence and vertical delineation suggests dominance as the discriminant factor.}
    \label{s:emotic_ValAr_scatter}
\end{figure}

To visualize how these emotions relate to each other, they were projected into the VAD space as shown in Fig. ~\ref{fig:circumplex}. In general, the categorical emotions share similar distributions for both datasets. However, looking at the plots for valence vs. dominance, there is a distinct cluster in the BoLD dataset for high-dominance, low-valence emotions such as ``Anger,'' ``Disapproval,'' ``Aversion,'' and ``Annoyance.'' 
Because these are all high arousal emotions, this difference between datasets may be caused by the additional motion information video samples have over static photos. 
Other than this cluster, there appears to be a strong correlation between dominance and valence. Past work has shown that these dimensions are not completely orthogonal~\cite{warriner_norms_2013}. The collected annotations from these two datasets seem to confirm this finding. Even with the distinct clustering off the mainline dominance/valence plot, this can be accounted for by their separation in the arousal dimension. The scatter plots in Fig.~\ref{s:bold_ValDom_scatter} show that the only categorical emotion outside of the ``Anger'' cluster that really benefits from the inclusion of dominance is confidence as its heatmap appears uniformly distributed across the valence axis while decreasing from high dominance to low dominance. 
Similar to the linear relationship between valence and dominance, in the raw valence-arousal projections in Fig.~\ref{fig:circumplex}, we also observe a collapse along the arousal dimension. This observation suggests changes in valence are the primary discriminative factor for differentiating emotions. This ``valence focus'' mirrors results from previous studies~\cite{feldman_valence_1995}. Still, when the labels from the BoLD dataset are projected onto the unit circle based on the mean valence and arousal for each emotion, the outcome broadly aligns with the affective circumplex~\cite{russell_circumplex_1980, barrett_structure_1999}.

In addition to the simple VAD embeddings (Fig. ~\ref{s:bold_DomAr_scatter}-\ref{s:emotic_ValAr_scatter}), we also once again produce UMAP embeddings~\cite{mcinnes_umap_2018} of the space as shown in Fig. \ref{fig:bold_umap}. We performed this analysis on both datasets. However, due to the large imbalance in the EMOTIC dataset as well a lack of data cleaning similar to BoLD's annotator reliability analysis, we limit the following analysis strictly to BoLD. First, it is important to note that although the dimensions are not meaningful, we can assign some meaning to them by also looking at the continuous VAD annotations in the same embedding space. In general, valence corresponds to the horizontal axis with arousal on the vertical. Similar to their correlation in the 2D VAD embeddings, dominance closely follows valence by generally increasing from left to right. For each individual categorical emotion, we then generate a heatmap so we can see where it exists in these embeddings. The discreteness of these heatmaps provides an idea of how fundamental each of these emotions is. For example, in Fig. 4 it is clear that ``Happiness'' and ``Anger'' occupy their own distinct clusters away from the main grouping. Likewise, Ekman's other basic emotions (Sadness, Fear, Surprise), also have relatively discrete clusters suggesting they are a fundamental building block for more complex emotions. On the opposite end of the spectrum, ``Anticipation'' and ``Engagement'' appear almost randomly dispersed throughout the embedding space. This spread suggests they are more complex in their expression and present themselves in a wider range of scenarios. 


\section{Discussion}\label{section:discussion}

Using word embeddings, we were able to show how existing models of human emotion popular in psychology either lack sufficient coverage or contain redundant, overlapping labels. Through agglomerative clustering, we were able to identify 15 components that minimized overlap and maximized coverage across cultures. This model, HICEM, was able to provide almost as much coverage with half as many labels as compared with existing emotion models. Interestingly enough, the only basic emotion described by Ekman which was not included in the HICEM was ``Disgust.'' The combination of ``Anger'' and ``Disgust'' in our analysis is similar to previous work on biologically basic emotions~\cite{jack_dynamic_2014}. 

Another consideration in the construction of comprehensive emotion models is the inclusion of general wellbeing/pain as well as a neutral affective state in the model. Although these are not classically thought of as emotions, the inclusion of these makes sense for HICEM because wellbeing/pain is expressed through facial expressions and body language similar to other emotions, and ``Neutral'' affective states also contain a variety of information useful for filtering out functional movements ({\it e.g.}, walking) common in everyday life. In addition to this, recent work taking inspiration from componential models of emotion has looked to relate other physiological systems to affective states. Hunger, thirst, sleepiness, and stress have all been shown to be connected to emotion~\cite{danziger_extraneous_2011, dancy_hybrid_2021}. These factors can be added to the components found in HICEM to provide complete coverage across both physiological processes and emotion. 

Although quantitative analysis shows diminishing returns after around 15 clusters, qualitatively the completeness of the model seems to peak between 25 and 30 emotion concepts similar to previous results in earlier studies~\cite{cowen_self-report_2017, cowen_mapping_2019}. Thus, it may be more beneficial to proceed with more than 15 base components in future dataset creation to ensure completeness. Likewise, if there are certain areas of the emotion space which are more important for a given domain, the hierarchical nature of our model easily allows for additional dimensions while maintaining maximum coverage. An example of this might be the inclusion of abnormal emotional states for mental health diagnoses. If we consider the clustering in Fig.~\ref{fig:en_clusters} for 15 components, instead of using ``bizarre'' as a label, we can look at the two clusters which formed it and split the label into ``zany'' and ``nonsensical.''

Finally, similar to previous work, it is recommended that HICEM is used in tandem with continuous affective dimensions ({\it i.e.}, VAD) to provide a holistic representation of the emotion space. Given the relationship between valence and arousal in existing large-scale datasets, it is recommended to drop dominance as a dimension in future data collection to reduce redundancies and costs associated with the data collection. Instead, alternate dimensions such as certainty or effort may be included. An interesting extension of HICEM would be to examine the latent continuous dimensions of emotion and develop a similar high-coverage model to describe the space. 

\subsection{Limitations} 

As stated previously, one limitation with using word embeddings has to do with antonyms being used in similar contexts, potentially having a higher cosine similarity than their synonyms. Although UMAP helps in increasing the distance between these antonyms in our analysis, this context issue still influences cluster purity because there are situations where it is difficult to tease out.
By taking the median vector for each cluster, we minimize this effect when generating our summary words. Another alternative would be to use word-level emotion distributions as described by Li et al.~\cite{li_word-level_2021} for text classification tasks. In addition to this, the use of Deep Learning (DL) generated embeddings~\cite{devlin_bert_2018} from free text annotations provides an interesting alternative. These do not rely on local word context to generate their vectors and have shown higher performance in NLP tasks compared to classical methods. 

In addition to the embeddings, our method also is limited by translation. We once again minimized the influence of this by averaging across several cultures; however, better results for each individual language may be achieved by having native speakers generate localized emotion word lists and then using those lists to fine-tune the models. Similarly, although we can generate a consistent set of labels for use across languages, the way people perceive these different emotion words is unique to their culture. Given that all languages appear to share the same 15 base components, transfer learning and fine-tuning can be performed to better fit machine learning models to their local language context.

\subsection{Applications} 

HICEM is primarily a descriptive tool, so its value comes from its ability to describe large numbers of affective states with relatively few labels. This is ideal for dataset annotation in modern data-driven AI because it maximizes the return on investment in terms of the amount of information gathered from each sample labeled. Although limited to 15 components, HICEM still provides high coverage over a wide range of human emotions.
This advancement means next-generation affective computing or AEI applications leveraging HICEM will allow for more natural human-machine or human-robot interactions. 

In addition to HICEM's value as an annotation tool, the process for developing our emotion model can also be used to build a taxonomy of human emotion similar to WordNet~\cite{miller_wordnet_1995}. Such a tool has potential applications in psychological research and can also be used for succinctly describing a patient's emotional state. Because HICEM is limited to discrete emotions, an interesting extension of this would be to use NLP word embeddings to find the equivalent for the continuous emotion dimensions.

\section{Conclusions}\label{section:conclusion}

There has been a substantial amount of research into computational techniques for recognizing human emotion. However, little work has been done examining the actual emotion models underpinning this research. As our analysis shows, existing models of emotion are insufficient for real-world applications. In affective computing, coverage is much more important than the completeness of the emotion model due to practical limitations in data collection and annotation. Here, using unsupervised techniques we were able to identify 15 components with minimal overlap and maximum coverage across 1,720 emotion concepts. This work provides a more efficient model of human emotion which is a step toward achieving artificial emotional intelligence.

\ifCLASSOPTIONcompsoc
  \section*{Acknowledgments}
\else
  \section*{Acknowledgment}
\fi

This research was supported by generous gifts from the Amazon Research Awards program. The project used computational resources from the Extreme Science and Engineering Discovery Environment (XSEDE), which is supported by National Science Foundation grant No. ACI-1548562.
The authors are grateful to Yu Luo for developing the BoLD dataset used in this study. The authors thank Tal Shafir and Reginald B. Adams, Jr. for helpful discussions. The authors appreciate the support and encouragement from Yelin Kim and Adam Fineberg.


%

\ifCLASSOPTIONcaptionsoff
  \newpage
\fi



%



\bibliographystyle{IEEEtran.bst}

\bibliography{IEEEabrv.bib,references.bib}

%

\begin{IEEEbiography}[{\includegraphics[width=1.1in,clip,keepaspectratio,trim=1 12 0 0]{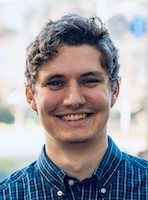}}]{Benjamin Wortman}
received the bachelor's degree in data sciences from The Pennsylvania State University, University Park, in 2020. He is expected to receive the MS degree in informatics from Penn State in 2022. He is also a PhD candidate in the Informatics program of the College of Information Sciences and Technology at Penn State. His research interests include affective computing, computer vision, and machine learning.
\end{IEEEbiography}


\begin{IEEEbiography}[{\includegraphics[width=1.1in,clip,keepaspectratio,trim=1 12 0 0]{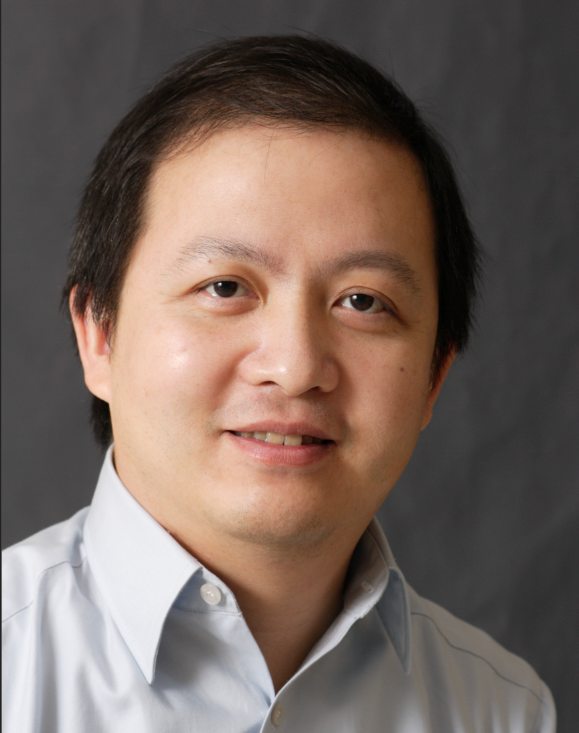}}]
{James Z. Wang}
is a Distinguished Professor of the Data Sciences and Artificial Intelligence section and the Human-Computer Interaction section of the College of Information Sciences and Technology at The Pennsylvania State University. He received the bachelor's degree in mathematics {\it summa cum laude} from the University of Minnesota (1994), and the MS degree in mathematics (1997), the MS degree in computer science (1997), and the PhD degree in medical information sciences (2000), all from Stanford University. His research interests include affective computing, image analysis, image modeling, image retrieval, and their applications. He was a visiting professor at the Robotics Institute at Carnegie Mellon University (2007-2008), a lead special section guest editor of the IEEE Transactions on Pattern Analysis and Machine Intelligence (2008), and a program manager at the Office of the Director of the National Science Foundation (2011-2012). He is on the editorial board of the IEEE BITS -- The Information Theory Magazine's special issue on Information Processing in Arts and Humanities (2022). He was a recipient of a National Science Foundation Career award (2004) and Amazon Research Awards (2018-2022).
\end{IEEEbiography}





\enlargethispage{-6in}

\end{document}